# Computational Curiosity

(A Book Draft)

by

Qiong Wu

wuqi0005@e.ntu.edu.sg

Nanyang Technological University

# Contents

# PREFACE

In recent years, researchers have shown an increasing interest in studying intelligent agents with various characteristics, such as affective agents, negotiation agents, trust agents, persuasive agents, and pedagogical agents. Each of the characteristics brings a new capability to intelligent agents and enhances certain aspect of their performances. These agents, however, still lack the capability to direct their attention towards novelty or to seek interestingness. In human beings, this capability is commonly observed, which is driven by the natural motivation: curiosity. Curiosity has been consistently recognized as the critical motivation that is associated with exploratory behaviors such as exploration, investigation, and learning. It has been identified as a driving force for child development, scientific research, and educational achievements. According to Kashdan, curiosity benefits human beings at two levels: the individual level and the social level. At the individual level, curiosity is associated with individual growth, as an ``innate love of learning without the lure of any profit''. At the social level, curiosity is an ingredient for enhancing interpersonal relationships, through infusing energy and passion into social interactions.

The many advantages of curiosity to human beings have inspired researchers to devise computational forms of curiosity to endow artificial beings (or agents) with desirable functions. For example, a curious design agent can arrange art exhibits to elicit the curiosity of their viewers and provide an aesthetically pleasing experience; a curious exploratory agent has been shown to achieve higher learning efficiency in unknown environments. From a machine learning perspective, curiosity has been proposed as algorithmic principles to focus learning on novel and learnable regularities in contrast to irregular noises. These algorithm principles make the agents to be ``explorative'' and allow them to have ``the desire to improve the model's knowledge about the world'', which have shown success in speeding up learning and building unsupervised developmental robotics.

Human beings, unfortunately, is often more complex than what a machine learning model can describe. Most machine learning algorithms are strictly utilitarian. The machine always remains enthusiastic to learn and mostly assumes that there is something to be learned. On the other hand, a person may behave less rationally. For example, a person may lack the will to learn even when ample opportunities for learning exist. Alternatively, a person may be curious about a stimulus but realize that little can be learnt from it due to a lack of knowledge or adverse environments (e.g. information blocked by a firewall). In order to provide a more complete model of human cognition and design artificial companions that can elicit curiosity (e.g. for pedagogical purposes), we believe it is beneficial to go back to the research in psychology to understand how human curiosity can be aroused.

Psychology studies suggest that curiosity can be externally stimulated by various factors, such as novelty, conflict, uncertainty, and complexity, etc. Each of these factors characterizes a different condition where curiosity can potentially be elicited. For example, novelty is induced by something new, whereas surprisingness occurs when an expectation based on previous knowledge is violated by the actual outcome. Several curiosity stimulating factors have been considered in computational curiosity and the explicit consideration of different curiosity stimulating factors makes the curious agents more human-like, more efficient in exploration, and more reactive in collaborative learning environments. Hence, we believe that the explicit consideration of various curiosity stimulating factors can enrich the current understanding of computational curiosity and open up new directions for future development.

In this book, we introduce a generic computational model of curiosity for intelligent agents based on the psychology of curiosity. This computational model of curiosity consists of abstract functions and their interactions between each other. Representing computational models for intelligent agents with abstract functions makes them general enough to allow different implementations in different application contexts. Based on this generic computational model of curiosity, we introduce a curiosity-driven learning algorithm and a curiosity-driven recommendation algorithm. The curiosity-driven learning algorithm showcases how curiosity benefits an agent at the individual development level, whereas the curiosity-driven recommendation algorithm demonstrates how curiosity benefits an agent at the social recommendation level. The curiosity-driven learning algorithm realizes the generic computational model of curiosity in a fast neural learning agent: the extreme learning machine, and is referred to as the Curious Extreme Learning Machine (C-ELM). Experimental comparisons with other popular classifiers on benchmark classification problems show a superior learning and generalization performance of C-ELM. In the context of social recommendation, a curiosity-driven recommendation algorithm is developed by realizing the generic computational model of curiosity in a recommender agent. Experimental studies with large scale real world data sets show that curiosity significantly enhances the recommendation coverage and diversity, while maintaining a sufficient level of accuracy.

To further evaluate the practical values of the generic computational model of curiosity, we discuss the study of it in the domain of virtual worlds for educational purposes. It has been shown in educational studies that curiosity is an important driving force for child development, scientific research, and educational achievements. Hence, it can be envisioned that considering the concept of curiosity in the design of virtual worlds may enrich their educational potentials. In this book, two types of pedagogical agents are chosen for study in virtual world based learning environments, a virtual peer learner and a virtual learning companion. The virtual peer learner is a Non-player Character that aims to provide a believable virtual environment without direct interactions with users, whereas the virtual learning companion directly interacts with users to enhance their learning experience. It can be expected that curiosity may allow the virtual peer learner to demonstrate more human-like learning behaviors in a virtual learning environment, and adds new ingredients into the interactions provided by the virtual learning companion.

In a word, this book discusses computational curiosity, from the psychology of curiosity to the computational models of curiosity, and then showcases several interesting applications of computational curiosity. A brief overview of the book is given as follows.

- Chapter 1 discusses the underpinnings of curiosity in human beings, including the major categories of curiosity, curiosity-related emotions and behaviors, and the benefits of curiosity.
- Chapter 2 reviews the arousal theories of curiosity in psychology and summarizes a general two-step process model for computational curiosity.
- Base on the perspective of the two-step process model, Chapter 3 reviews and analyzes some of the traditional computational models of curiosity.
- In Chapter 4, we introduce a novel generic computational model of curiosity, which is developed based on the arousal theories of curiosity. This computational model of curiosity consists of abstract functions and their interactions between each other, and such a representation method allows different implementations in different application contexts.
- After the discussion of computational models of curiosity, we outline the important applications where computational curiosity may bring significant impacts in Chapter 5
- Chapter 6 discusses an implementation of the generic computational model of curiosity in a machine learning algorithm. This algorithm realizes the generic computational model of curiosity in an extreme learning machine based classifier, which is referred to as the Curious Extreme Learning Machine classifier (C-ELM). The performance of C-ELM is evaluated against other popular classifiers in the literature on benchmark data sets.
- Chapter 7 discusses an implementation of the generic computational model of curiosity in a recommender system. This curiosity-driven recommender system realizes the generic computational model of curiosity in the popular matrix factorization algorithm, which largely enhances recommendation diversity and coverage. The performance of the curiosity-driven recommendation algorithm is evaluated using two large scale real world datasets.
- In Chapter 8 and Chapter 9 we study of the generic computational model of curiosity in two types of pedagogical agents. In Chapter 8 a curious peer learner is developed. It is a non-player character that aims to provide a believable virtual learning environment for users. The effects brought by curiosity to virtual peer learners are studied through computer simulations. In 9 a curious learning companion is developed. It aims to enhance users' learning experience through providing meaningful interactions with them. The curious learning companion realizes the generic computational model of curiosity and is carefully evaluated with human users through field studies.
- Chapter 10 discusses the open questions in the research field of computation curiosity.

# CHAPTER 1 PSYCHOLOGY UNDERPINNINGS OF CURIOSITY

In human beings, curiosity is closely related to cognition, emotion and behavior. It underlines human cognitive development, aesthetic appreciation, and interpersonal relationships. In this chapter, we review the literature in psychology on the underpinnings of curiosity and the benefits of curiosity for human beings.

## 1.1. Categories of Curiosity

Most psychologists believe that curiosity is an intrinsic motivation driving the cognitive development of both humans and animals alike. Berlyne [Berlyne (1960a)] categorized curiosity along two spectrums: (1) from perceptual curiosity to epistemic curiosity, and (2) from specific curiosity to diversive curiosity. Perceptual curiosity, which resides in the lower level of cognition, stems from the senses of both animals and humans (e.g. senses of touch, vision, taste, etc.). It is defined as "a drive that is aroused by novel stimuli and reduced by continued exposure to these stimuli" [Loewenstein (1994)]. Epistemic curiosity, referred to as "an appetite for knowledge", is related to the higher level of cognition and believed to be a distinctive human feature. While perceptual curiosity and epistemic curiosity are defined along the lines of "lower" and "higher" levels of cognition, specific curiosity and diversive curiosity are distinguished by the possibility of curiosity having a "direction". Specific curiosity is aroused by a particular piece of information. Diversive curiosity is a general drive to seek information with no specific direction and is predominantly employed to relieve boredom.

Berlyne's cognitive account for curiosity has been a theoretical foundation for the majority of recent studies. Litman and Speilberger [Litman and Speilberger (2003)] agreed with Berlyne that there is a salient difference between diversive curiosity and specific curiosity, and conducted experimental analysis to provide scales for measuring both concepts. They further concluded that diversive curiosity and specific curiosity, as well as perceptual curiosity and epistemic curiosity, are "substantially correlated". Speilberger and Starr [Speilberger and Starr (1994)] associated diversive curiosity with "low-level" stimuli and specific curiosity with "high-level" stimuli. However, Schmitt and Lahroodi [Schmitt and Lahroodi (2008)] disagreed with the notion that curiosity can be diversive. They argued that curiosity can only be specific towards its possessor and objects that cause the curiosity. Instead of diversive curiosity, they referred to the generic desire for knowledge as "inquisitiveness".

Nevertheless, a general consensus among the psychologists points to the close relationship between curiosity and cognition, as a drive to explore novel stimuli or an appetite for knowledge.

## 1.2. Curiosity-Related Emotions

Curiosity is related to emotional constructs. In an early account of curiosity by James [James (1950)], curiosity is viewed as an instinctual or emotional response closely related to fear. He believed that curiosity motivates organisms to actively explore their environment, whereas fear tends to turn the organisms away from the risks induced by unfamiliarity. Berlyne's branch of research, referred to as "drive theory" by Loewenstein [Loewenstein (1994)], is based on the assumption that curiosity produces an unpleasant sensation that is reduced by exploratory behaviors. Loewenstein believed that rather than serving a purposive end, the primary objective of satisfying one's curiosity is to induce pleasure.

Wundt [Wundt (1874)] introduced the theory of "optimal level of stimulation", which serves as a general rule postulating the relationships between stimulus intensity and the hedonic tone. Based on this theory, Berlyne [Berlyne (1960b)] proposed that there is a need of "intermediate arousal potential" for curiosity to be aroused. Berlyne's theory demonstrates that too little stimulation can result in boredom or inaction, while too much stimulation may result in aversion or withdrawal. Only when the level of stimulation is optimal and pleasurable can exploratory behaviors occur.

From the above discussion, it can be seen that curiosity is closely related to emotional constructs such as fear, pleasure, boredom and anxiety. The decision on whether to explore or to avoid a stimulus is driven by emotional comfort and results in behaviors that regulate emotional states.

## 1.3. Curiosity-Related Behaviors

The most salient expression of curiosity is through exploratory behaviors, by which curiosity can be satisfied. Berlyne [Berlyne (1960b)] defined two levels of exploratory behaviors, one associated with the perceptual level of curiosity, and the other associated with the epistemic level of curiosity. At each level, the exploratory behaviors can take many forms. At the perceptual level of curiosity, Berlyne divided exploratory behaviors into three categories according to the nature of responses. He referred to the exploratory behaviors as orienting responses if they consist of changes in posture, orientations of sensory organs, or states of sensory organs. The second category of exploratory behaviors is associated with locomotion, such as approaching or withdrawing from the stimuli. When an exploratory behavior causes changes in external objects, through manipulation or otherwise, it is called an investigatory response.

At the epistemic level of curiosity, Berlyne also defined three categories of exploratory behaviors. The first category is observation, which places the subject in contact with external situations that can nourish the pertinent learning process. The second category is thinking, which refers to "productive" and "creative" thinking, rather than "reproductive thinking" that only calls up memories to determine how problems should be handled. The last category is consultation, which exposes an individual to verbal stimuli issued from other individuals, including questioning and reading, etc.

In summary, through exploratory behaviors, cognitive growth is achieved by creative thinking and emotional states are regulated towards a feeling of "pleasure".

## 1.4. Benefits of Curiosity

Current research indicates that curiosity can contribute to human well-beings at two distinct levels: individual level and social level. At the individual level, curiosity provides an emotional motivation for self-development. It is the driving force for child development as well as an important spur for educational achievements [Loewenstein (1994)]. Also, literature in psychology indicates a close relationship between curiosity and aesthetics, humor, and fun. According to Berlyne [Berlyne (1960b)], these behaviors have common motivational factors and are reinforced by common sources of rewards.

At the social level, curiosity can enhance interpersonal relationships. By studying the role of curiosity in conversations, Kashdan et al. [Kashdan et al. (2011)] suggested that curiosity can build social bonds by promoting behaviors such as engagement, responsiveness, and flexibility. These are desirable behaviors for developing interpersonal relationships and building intimacy. Their findings indicate that curiosity is uniquely related to the development of interpersonal closeness with strangers [Kashdan et al. (2004)].

# CHAPTER 2 AROUSAL THEORY

In this chapter, we review the literature in psychology on the arousal mechanisms of curiosity. This review will provide insights into possible ways of implementing curiosity in artificial beings and is the basis for conducting our research.

According to Loewenstein [Loewenstein (1994)], existing psychological theories on human curiosity can be divided into three categories: incongruity theory, competence theory, and drive theory. Incongruity theory holds on to the idea that curiosity is evoked by the violation of expectations [Hebb (1949)]. Competence theory views curiosity as an intrinsic motivation to master one's environments [White (1959)]. Drive theory believes in the existence of a curious drive, either primary (homeostatic generated in a similar way as hunger) or secondary (externally generated by stimuli). As commented by Loewenstein, the incongruity theory and the competence theory suffer from the same deficiency that both fail to offer a comprehensive account of curiosity. Hence, in this work, we focus on the drive theory and adopt Berlyne's interpretation of curiosity.

According to Berlyne [Berlyne (1960b)], traditional psychological research concentrated on problems of response selection, which studies what response humans will make to one standard stimulus at a time. However, curiosity deals with a different problem from response selection, which is referred to as stimulus selection. Stimulus selection discusses when several conspicuous stimuli are introduced at once, to which stimulus humans will respond. Berlyne has conducted extensive experimental studies to understand the process of stimulus selection and discovered a set of collative variables that govern this process.

## 2.1. Collative Variables

Collative variables, according to Berlyne [Berlyne (1960b)], refer to the external factors that govern various forms of stimulus selection. There are four major collative variables, viz., novelty, uncertainty, conflict and complexity. Berlyne named these factors collative variables because the evaluation of each variable involves an analysis of similarities and differences between elements in a stimulus pattern. In the following part of this section, the major collative variables are reviewed.

**Novelty** denotes something new. In accordance with the time when a stimulus has been experienced by an organism, novelty can be divided into short-term, long-term, and complete novelty. A stimulus can either be completely new to an organism, or it could have been experienced within just the last few minutes. In the former case, it is called complete novelty, and in the latter case, it is called short-term novelty. The intermediate case where a stimulus has not been experienced for a period of time (usually days) is called long-term novelty. Based on whether a stimulus possesses qualities that have been perceived before, it is classified into absolute novelty or relative novelty. If the stimulus does not have any previously perceived quality, then it is absolute novelty; otherwise, it is relative novelty.

Based on these observations, Berlyne introduced three criteria to measure novelty, i.e., novelty is inversely related to (1) how often the stimuli have been experienced before, (2) how recently the stimuli have been experienced, and (3) how similar the stimuli are to previously experienced ones.

Novelty is often accompanied by other properties, each of which may have different influences on exploratory behaviors. Berlyne listed them as supplementary variables to novelty, which includes change, surprisingness, and incongruity. A **change** of the stimulus in question may induce some priority in exploratory directions. **Surprisingness** arises when there is a stimulus that induces an expectation and a later stimulus that contradicts the expectation. **Incongruity** is somewhat different from surprisingness (e.g., the statement that the Earth is round is a surprise to people who are accustomed to the concept that the Earth is _at), and indicates an expectation that is not met by the same stimulus (e.g., a person who is used to receiving gifts on his birthday before may find the experience incongruent when nobody sends him birthday gifts this year).

**Uncertainty** arises when an organism has difficulty selecting a response to a stimulus. Berlyne adopted information theory to quantify uncertainty. He proposed to measure the uncertainty caused by a stimulus with the following steps: (1) draw up a list of stimuli that might occur (as a response to the stimulus in question), (2) partition them into classes, and (3) assign a probability to each class. The probability $p_i$ of each class denotes the competing strength of each possible response, and Shannon's entropy $H = -\sum_{i=1}^{n}(p_i \times log_2(p_i))$ denotes the degree of uncertainty.

**Conflict** occurs when a stimulus arouses two or more incompatible responses in an organism. A response can be incompatible with one another due to different reasons. Firstly, some responses may be innately antagonist to each other. For example, no organism can move forward and backward at the same time. Secondly, some responses may initially be capable of performing together but become incompatible through learning. For example, we seldom frown when shaking hands. Another reason for incompatible responses may be attributed to the limitation of an organism's ability to multi-task. For example, it would be considered an outstanding ability if a person can read two books at the same time. Berlyne proposed four criteria for measuring conflict. Conflict is positively related to (1) the nearness to equality in the strengths of competing responses, (2) the absolute strengths of competing responses, (3) the number of competing responses, and (4) the degree of incompatibility between competing responses.

**Complexity** roughly refers to the variety or diversity in a stimulus pattern. Three most obvious properties that determine the degree of complexity are: (1) the number of distinguishable elements in a stimulus, (2) the dissimilarity between these elements, and (3) the degree to which several elements are perceived and responded to as a unit.

In summary, collative variables are properties of a stimulus that describe the curiosity stimulating conditions. According to Berlyne, they are all eminently quantitative properties that exist in varying degrees. This endows them potential candidates for measuring the stimulation level of curiosity. Most of the computational models of curiosity have considered at least one of the collative variables for the determination of stimulation level.

## 2.2. Intermediate Arousal Potential

The existence of a stimulus does not necessarily result in curiosity. The arousal of curiosity depends on an appropriate level of stimulation induced by a stimulus. In the 1870s, Wundt [Wundt (1874)] introduced the concept of "optimal level of stimulation" and postulated an inverted U-shape relationship between the stimulation level and the hedonic value, referred to as the Wundt curve (Figure 2.1). It is a general rule stating that many forms of stimulation are pleasant at medium intensities and become unpleasant when their intensities are too high. Based on Wundt's theory and other experimental results from the literature, Berlyne formed the theory of "intermediate arousal potential", where too little stimulation results in boredom, too much stimulation results in anxiety and only intermediate stimulation results in curiosity. The two ends of the spectrum in the Wundt curve reflect two rules in stimulus selection: Avoidance of Boredom (AoB) and Avoidance of Anxiety (AoA).

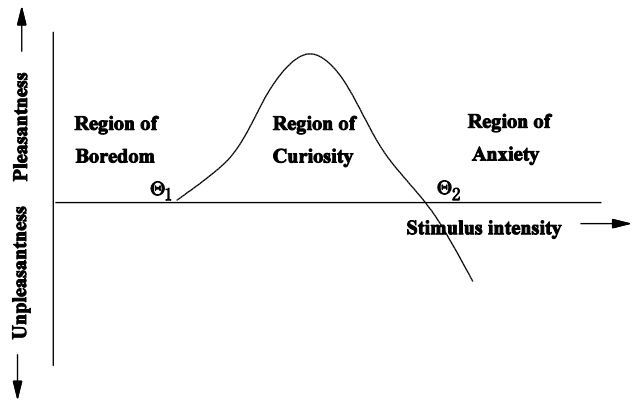

**Figure 2.1 The Wundt Curve (adopted from [Berlyne 1960b])**

To summarize, from the psychological point of view, the arousal process of curiosity can be abstracted into a two-step process model, which offers a uniform angle for examining and analyzing the computational models of curiosity:

- Step 1: evaluation of the stimulation level based on collative variables.
- Step 2: evaluation of the interest level based on the principle of intermediate arousal potential.

# CHAPTER 3 TRADITIONAL COMPUTAIONAL MODELS OF CURIOSITY

In this chapter, we review and analyze the traditional computational models of curiosity from the perspective of the proposed two-step process model summarized at the end of the previous chapter.

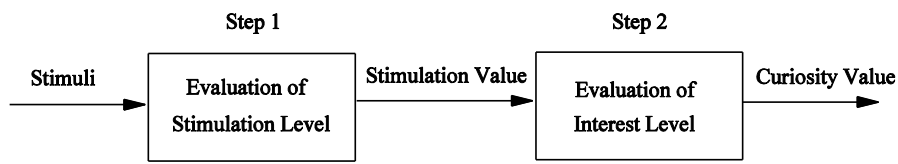

Figure 3.1 A general appraisal process of computational curiosity.

Based on the two-step process model, a general appraisal process for computational curiosity can be illustrated as in Figure 3.1. In this figure, the input stimuli are data samples perceived by the agent. The input stimuli trigger the agent's computational appraisal process of curiosity. In step 1, the appraisal process first evaluates the level of stimulation elicited by the stimuli, based on collative variables. Some of the existing models adopt a single collative variable to determine the stimulation value (e.g., [Saunders and Gero (2001)]), while others aggregate multiple collative variables to derive the stimulation value (e.g., [Macedo and Cardoso (2005)]).

In step 2, the level of curiosity is evaluated through a mapping from the stimulation value to the curiosity value. Some of the existing models follow the principle of "intermediate arousal potential" by explicitly simulating the Wundt curve, which represents a nonlinear mapping from the stimulation value to the curiosity value (e.g., [Saunders and Gero (2001); Merrick et al. (2008)]). These models accommodate both AoB and AoA in their stimulus selection approaches. In comparison, other models simply use the stimulation value as the curiosity value (e.g., [Schmidhuber (1991b); Oudeyer and Kaplan (2004); Macedo and Cardoso (2001)]). Some of these models also consider the principle of "intermediate arousal potential" when determining the stimulation value (e.g., [Schmidhuber (1991b); Oudeyer and Kaplan (2004)]), which accommodate both AoB and AoA. The rest of these models simply assume a positive correlation between the stimulation value and the curiosity value, which only support AoB but not AoA. In this case, a high stimulation can lead the agent to anxiety (e.g., [Macedo and Cardoso (2001)]).

Taxonomy for the existing computational models of curiosity is provided in Table 3.1. The models are classified into five categories according to the collative variables used for the evaluation of stimulation in Step 1. The five categories include novelty, surprise, uncertainty, change, and complexity. Within each category, the models are further classifies into two sub-categories, according to whether the principle of "intermediate arousal potential" is followed in the models. The first sub-category follows this principle and supports both AoB and AoA in stimulus selection, while the second sub-category assumes a positive correlation between the stimulation value and the curiosity value, which only supports AoB. Next, we will discuss each of the models following this taxonomy

**Table 3.1 Taxonomy of the existing computational models of curiosity in literature**

| Collative Variables | Avoidance of Boredom & Avoidance of Anxiety | | Avoidance of Boredom | |
|---|---|---|---|---|
| Novelty | Similarity & Stimulation of the Wundt curve | Saunders & Gero [01] Maher et al. [08] | Similarity | Macedo & Cardoso [99] |
| | | | Frequency & Similarity | Ogino et al. [06] |
| Surprise | Prediction error In zero sum games | Schmidhuber [99] | Prediction error | Schmidhuber [91a] Uğur et al. [07] |

|  | Information gain | Storck et al. [95] | Improbability | Macedo & Cardoso [99] |
|---|---|---|---|---|
| Uncertainty | - |  | Entropy | Macedo & Cardoso [95] |
| Change | Spread adjustment in Gaussian | Karaoguz et al. [11] | - |  |
| Complexity | Prediction improvement between successive situations | Schmidhuber [91b] | - |  |
|  | Prediction improvement between similar situations | Oudeyer & Kaplan [04] |  |  |
|  | Compression improvement | Schmidhuber [06] |  |  |
|  | Discriminability difference between two spaces | Pang et al. [09] |  |  |

## 3.1. Models based on Novelty

To explore the possibility of artificial creativity, Sanders and Gero [Saunders and Gero (2001)] developed a computational model of curiosity for intelligent design agents, focusing on the appraisal of novelty. The goal of such an agent is to evaluate the interestingness of creative design patterns based on individual experiences, where curiosity is the key for the evaluation of interestingness [Saunders and Gero (2001)]. In their model, each creative design pattern (generated by a design generator) can be considered as a stimulus and the stimuli experienced in the past are stored in a "conceptual design space". This conceptual design space is modeled by a Self-Organizing Map (SOM), representing the forms that the agent has experienced often enough to learn.

If we view their model from the perspective of the two-step process model, in Step 1, the evaluation of stimulation is governed by the degree of novelty in a stimulus. The evaluation of novelty is achieved by comparing the stimulus (a creative design pattern) with past experiences (the SOM representations), and is defined by the complement of the typicality measure. This implementation is in line with Berlyne's third criteria for measuring novelty: novelty is inversely proportional to similarity (typicality).

In Step 2, Saunders and Gero adopted the principle of intermediate arousal potential and explicitly modeled the Wundt curve as a non-linear function. This function is defined by the difference of two sigmoid functions as follows [Saunders (2002)]:

$$H = R - P$$

$$R = \frac{R_{max}}{1 + e^{-\rho_R(novelty - R_{min})}}$$

$$P = \frac{P_{max}}{1 + e^{-\rho_P(novelty - P_{min})}}$$

where $H$ is the curiosity value, $R$, P, $R_{max}$, $P_{max}$, $\rho_R$, $\rho_P$, $R_{min}$ and $P_{min}$ are the reward, punishment, maximum reward, maximum punishment, slope of the reward sigmoid functions, slope of the punishment sigmoid functions, minimum novelty to be rewarded and minimum novelty to be punished, respectively. This curiosity value is the evaluation of the interestingness of a design pattern or how curious it is about the design pattern.

Later, Merrick and Maher applied this model of curiosity in Non-Player Characters (NPCs) and reconfigurable robots [Merrick et al. (2008); Merrick and Maher (2009); Maher et al. (2008); Merrick and Huntington (2008); Merrick (2008a)] to enhance their performances. For NPCs, the goal of infusing curiosity is to achieve creative behaviors, namely, behavioral diversity [Merrick et al. (2008); Maher et al. (2008)]. These works are rooted in a motivated reinforcement learning framework, by which NPCs can learn the state-action policy in a virtual environment. Here, each event is regarded as a stimulus and is defined as the change between the current state and the previous one. All previously experienced events are stored in an SOM structure. A habituation layer of neurons is connected to the clustering layer and a habituation function is used to compute novelty. The final curiosity value is obtained by feeding the novelty value into the simulated Wundt curve (Equation 3.1). The curiosity value is used as an intrinsic reward to update the policy that maps states to actions. This curiosity reward can eventually lead to creative behaviors, because each new interesting event results in a perturbation of the existing behavioral patterns and form reward signals for the learning process.

For reconfigurable robots [Merrick and Huntington (2008)], the role of curiosity is to direct the robots' attention to reconfigurations in their structures. This is an important skill for them to learn new behaviors in response to structural changes. For robots, an event can be considered as a stimulus, which is defined as the change between the current sensation and the previous one. Following a similar algorithm in curious NPCs, curiosity rewards are generated for reconfigurable robots. With the curiosity rewards, the robots are self-motivated to explore changes in their structures and develop new behaviors.

In the domain of planetary exploration and map-building of interiors, Macedo and Cardoso [Macedo and Cardoso (1999)] proposed a model of curiosity for intelligent agents to simulate human-like exploratory behaviors in unknown environments. Along their research, they gradually introduced novelty, surprise, and uncertainty into their computational model of curiosity. Here, we first look at their model of novelty. Macedo and Cardoso's model relies on graph-based mental representations of objects. Each object can be regarded as a stimulus. In Step 1, the level of novelty regarding a stimulus (object) is measured based on the error correcting code theory of Hamming. Three steps are considered: 1) representing each graph (describing an object) in a common shape matrix 2) extracting the numerical code from the matrix representation of each graph, and 3) computing the Hamming Distance. Here, novelty is defined as the minimum Hamming Distance from the stimulus to the ones that have been experienced before, which determines the stimulation value. In Step 2, the stimulation value is directly used as the curiosity value. The system only supports AoB because the chance of an object to be explored is positively correlated to the level of curiosity, which direct the system away from boredom.

For both the models proposed by Sanders & Gero [Saunders and Gero (2001)] and Macedo & Cardoso [Macedo and Cardoso (1999)], regardless of the implementation, i.e., atypicality in SOM or hamming distance in graph-based mental representations, novelty reflects a comparison between the current stimuli and previous experiences. In other words, similarity. In some later models of curiosity, the factor of time, as another dimension in addition to similarity, is considered for measuring novelty.

Ogino et al. [Ogino et al. (2006)] addressed lexical acquisition problems for robots using computational curiosity to associate visual features of observed objects with the labels (for objects) that are uttered by a caregiver. In their work, each object can be interpreted as a stimulus for the robot. Visual features of each object are represented by an SOM. In Step 1, stimulation value is determined by novelty. The novelty of each object is calculated based on two types of saliency: habituation saliency, reflecting the temporal infrequency; and knowledge-driven saliency, reflecting the level of dissimilarity. The habituation saliency is characterized by habituation and inversely related to the frequency of the observation of a visual feature.

The knowledge-driven saliency is characterized by the acquired knowledge, where more saliency is given to visual features that are not associated with other labels. The product of the two saliency values represents the overall level of stimulation, which is directed used as curiosity value (in Step 2). The robot chooses to learn about the object with maximum curiosity value, which drives the system away from boredom and considers the AoB rule in stimulus selection. Their experimental results show that the infusion of curiosity helps accelerate the robot's learning.

## 3.2. Models based on Surprise

Two interpretations for surprise exist in the literature of computational curiosity. The first one interprets surprise as the difference between an expectation and the real outcome. Prediction error matches well with this interpretation and has been utilized in many curiosity models to measure the level of surprise [Schmidhuber (1991a); Barto et al. (2004a); Schmidhuber (1999); Uğur et al. (2007)]. The second interpretation describes surprise as the degree of not expecting something. Storck et al. [Storck et al. (1995)] modeled this type of surprise using the information gain before and after an observation, while Macedo and Cardoso [Macedo and Cardoso (2001)] proposed another measure using improbability. Next, we will discuss each of these models in detail.

*Prediction Error-based Models*: Schmidhuber [Schmidhuber (1991a)] introduced artificial curiosity into model building control systems. The goal of such a control system is to learn the input-output mapping in a noisy environment. Curiosity is infused to give the control system an intrinsic desire to improve the model's knowledge about the world. This is realized by introducing an additional reinforcement unit on top of the controller, which rewards actions that cause high prediction errors. The prediction error inherently measures the degree of mismatch between belief and reality, which is in line with the first interpretation of surprise. In Step 1, the stimulation value is determined by surprise (prediction error) and it is directly used as the curiosity value (in Step 2) to form decisions (on rewarding the system). This mechanism (rewarding the system with high surprise/curiosity) encourages certain past actions to be carried out again in order to repeat situations similar to the mismatched ones. Hence, the system will always direct its attention toward something that is unknown and therefore avoid boredom. This working mechanism only supports AoB in stimulus selection. Barto's theory of intrinsically motivated reinforcement learning is implemented based on a similar principle [Barto et al. (2004a); Singh et al. (2004)].

In a later variation, Schmidhuber [Schmidhuber (1999, 2002)] worked on exploring the space of general algorithms that can automatically create predictable internal abstractions of complex spatial-temporal events. In this model, both AoB and AoA are accommodated. Curiosity is interpreted as the ability of the system to focus on interesting things by losing interest in the overly predictable (boring) or the overly unpredictable (anxiety inducing) aspects of the world. To achieve this goal, the system is realized by two intrinsically motivated agents playing zero-sum games. The two agents can bet in advance on whether a prediction is true or false. If they bet on different outcomes, the system will check who is right. The winner gets rewarded by receiving the other's bids whereas the loser loses its bids due to surprise (error in prediction). Hence, both agents are motivated to lure the opponent into agreeing on computation sequences that will surprise the other one. However, a surprised module will eventually adapt and in turn, cause the other agent to lose a source of reward. In this way, the system as a whole is motivated to shift the exploration focus and reveal the unknown yet predictable regularities.

Another example of prediction error-based implementation of surprise is given by the work of Uğur et al. [Uğur et al. (2007)]. In situations where a robot physically interacts with the environment to explore and learn, assuming the traditional reinforcement learning method is applied, the robot may require a large number of interactions to learn even simple tasks. This could eventually damage the robot. To address this problem, Uğur adopted a Support Vector Machine (SVM) to learn the perception and action mapping in the robot, where curiosity is introduced to select interesting training data for SVM to reduce the total number of training data required. In Step 1, the stimulation level is determined by surprise, which is measured by a sample's distance to the hyper-plane (which separates two classes) in the feature space. In Step 2, the stimulation value is directly used as curiosity value for decision making. The system supports AoB because it can be driven away from boredom: only if the distance is smaller than a fixed threshold (curiosity value is high), the sample is considered interesting and sent for learning.

*Probability-based Models*: An extension of Schmidhuber's curiosity principle to non-deterministic environments was done by Storck et al. [Storck et al. (1995)]. The goal of their system is to learn the model of a nondeterministic Markov environment where each state-action pair $(S(t); a(t))$ may result in different next states $(S(t + 1))$ probabilistically. The goal of introducing curiosity into this learning system is to actively search for interesting training examples that can maximize expected learning improvement. In step 1, the stimulation value is determined by surprise, which reflects "the degree of not expecting something". They adopt a probabilistic way of measuring surprise, which is the Kullback-Leibler distance between the agent's belief distribution before and after making an observation. This value of information gain reflects the learnability of the exploration space. Exploration areas where little information gain can be achieved are either too predictable (well-learnt) or too unpredictable (inherently unlearnable). Hence, the information gain (surprise) accommodates both AoB and AoA in stimulus selection. In step 2, the surprise value is directly used as the curiosity value, which acts as rewards for the reinforcement learning of state-action pairs. In this way, curiosity drives the system away from the exploration space where little information gain can be achieved, i.e., areas that are either too boring or anxiety-inducing.

Macedo and Cardoso [Macedo and Cardoso (1999, 2001, 2004, 2005)], mentioned in Section 3.1, also modeled human forms of surprise in artificial agents. Surprise serves as an intrinsic drive to direct the agent's exploration in unknown environment. Their model relies on graph-based mental representations of objects, where each object is considered as a stimulus. The surprise level is defined by the degree of not expecting a stimulus (object), and is implemented as the improbability of existence of the stimulus. However, surprise in this model is not treated as a stimulating factor for curiosity. Hence, the surprise value is not mapped to the curiosity value.

### 3.3. Models based on Change

In robot sensory-motor learning, Karaoguz et al. [Karaoguz et al. (2011)] proposed a model of curiosity to direct robots' attention to changes in the environment. The aim of the model is not to measure the amount (or level) of changes in a stimulus, but rather to focus on how to react when changes occur. Hence, the system does not provide evaluation for the level of stimulation or the level of curiosity (induced by change), but offers mechanisms to redirect its attention to changes.

For the robot, the focus of attention is determined by a Gaussian distribution that governs the sampling of training examples over the laser motor space. The center of the Gaussian distribution is determined by the mean value of the last N samples added to the mapping, and directs the attention of the robot to areas where new samples have been recently added. The spread of Gaussian distribution is related to the performance (success rate) of the mapping by

$$\sigma(t, s_r) = \sigma_0 + k_b t_b + k_f (t_f - s_r)^2$$

where $\sigma_0$ is the baseline, $t_b$ is the number of time steps since the last new sample was added, $k_b$ is the boredom coefficient, $t_f$ is the failure threshold, $s_r$ is the success rate, and $k_f$ is the failure coefficient.

It can be seen from Equation 3.3 that the attention spread is inversely correlated to the success rate. For example, the attention spread is wide when the success rate is low due to a high value of $k_f(t_f - s_r)^2$. When a change happens, $t_b$ will be reset to 0 and the high past success rate of that region will result in a narrow Gaussian distribution of samples. Newly-added links in unlearnable areas (noisy region) will fail on retest, which can drive the success rate for that region below the failure threshold $t_f$ and increase the width of sampling distribution, redirecting the system away from that region. Hence, the system has a drive towards regions where the existing mappings are not making accurate predictions (AoB) and a drive away from the regions where no improvements can be obtained (AoA).

From the perspective of machine learning, one can argue that this model is a special case of the earlier introduced models based on predictors [Schmidhuber (1991a,b)], because a predictor always predicts that the next observation is similar to the current observation, which indicates no change. Once a change occurs, the predictor makes an error and updates its knowledge about the world.

### 3.4. Models based on Uncertainty

Uncertainty arises when there is no clear response to a stimulus [Berlyne (1960b)]. The entropy in information theory has been proposed to measure the degree of uncertainty in a stimulus. This measure has also been employed very often in computational implementations to realize uncertainty-based curiosity [Macedo and Cardoso (2005)]. To improve an agent's exploration in unknown environments, Macedo and Cardoso [Macedo and Cardoso (2005)] introduced a measure of uncertainty, on top of novelty [Macedo and Cardoso (1999, 2001, 2004)], into their model of curiosity. Based on the same system setup as presented in Section 3.1, Macedo and Cardoso [Macedo and Cardoso (2005)] argued that the desire to know or learn an object can be induced by both novelty and uncertainty. Each object (stimulus) can contain known parts (without uncertainty) and uncertain parts. The known parts of an object are used to measure novelty through Hamming Distance(introduced in Section 3.1), whereas uncertainty is measured by the entropy of all uncertain parts, including analogical and propositional descriptions of the physical structure, and functions of the object. In step 1, the stimulation value is determined by the aggregation of novelty and uncertainty. In step 2, the stimulation value is directly used as curiosity value. The system adopts AoB in stimulus selection and chooses objects with highest curiosity value to explore.

## 3.5. Models based on Complexity

In machine learning models of curiosity, complexity has been associated with the predictive power of predictive systems or compressive power of data compression systems [Li and Vitanyi (2008); Schmidhuber (1991b, 2006); Oudeyer and Kaplan (2004)]. In this subsection, we review these models and their variations.

In Section 3.2, we introduced Schmidhuber's early ideas on implementing curiosity by rewarding the system proportionally to surprise, i.e., prediction error [Schmidhuber (1991a)]. However, this implementation only succeeds in guiding the system to avoid boredom (i.e., well learnt area) but not anxiety (i.e., inherently unpredictable area caused by noisy). Later, Schmidhuber refined the model to accommodate both AoB and AoA. He defined curiosity as a simple principle: to learn a mapping from actions to the expectation of future performance improvement [Schmidhuber (1991b,c)]. Instead of pure prediction error, reward is now generated according to the controller's prediction improvement, i.e., change in prediction error. In step 1, the complexity of a data to be learnt by the system (e.g., familiar data that is easy to learn or noises that are too difficult to learn) is determined by prediction improvement, which supports both AoB and AoA in stimulus selection. The prediction improvement is obtained by a confidence module, which evaluates the reliability of a prediction and can be realized by probability-based or error-based methods. In step 2, the stimulation value is directly used as curiosity value, which is adopted as intrinsic rewards for learning. In this way, the controller will choose actions (based on the delayed reward) to deliberately select training examples with easily learnable regularities. Hence, the infusion of curiosity can direct the system's attention away from the exploration space that is either too predictable or too unpredictable.

Schmidhuber [Schmidhuber (2006, 2009a,b)] formed his formal theory of creativity, by generalizing the simple principle of curiosity from predictors to data compressors. According to this theory, a 'beautiful' sensory data is one that is simple yet has not been fully assimilated by the adaptive observer, which is still learning to compress data better. The agent's goal is to create action sequences that can extend the observation history to yield previously unpredictable but quickly learnable algorithmic regularities. In other words, it is looking for data with high compressibility (reflected by curiosity value).

Schmidhuber's implementation of prediction improvement by nature is a comparison of prediction error between situations that are successive in time. This principle allows robots to avoid long periods of time in front of a television with white noise (completely unlearnable situations) because the prediction error will remain large and the robot will be bored due to little prediction improvement. However, this principle is not robust enough in the alternation of completely predictable and unpredictable situations, because robots can get stuck here due to large prediction improvements. To cope with such problems, Oudeyer and Kaplan [Oudeyer and Kaplan (2004); Oudeyer et al. (2005)] refined Schmidhuber's simple principle. Instead of comparing the prediction error between situations that are successive in time, they compare the prediction error between situations that are similar. They proposed a model of curiosity that allows a robot to group similar situations into regions where comparison between situations is meaningful. The learning space is divided into regions and each region has an expert to make local predictions. Each expert computes the prediction improvement (curiosity value) locally and rewards its state-action pairs according to the prediction improvement. This works well in practice to handle problems such as robots get stuck in situations with completely predictable and unpredictable sample data in alternation.

Another variation of using prediction improvement as curiosity drive has been proposed by Pang et al. [Pang et al. (2009)]. This model is rooted in incremental Linear Discriminant Analysis (LDA). Here, curiosity is measured as the discriminability difference (residue) between the LDA transformed space and the original space. The infusion of curiosity can help the system actively search for informative examples to learn and improve the performance using fewer instances.

## 3.6. Discussion

An interesting point worth noting is that, from the machine learning perspective, there are certain recurring principles underlying curiosity-inspired algorithms. The first group of recurring principles includes generating intrinsic curious rewards based on errors [Schmidhuber (1991a); Uğur et al. (2007)] or Shannon's information [Scott and Markovitch (1989)]. This group of principles can redirect learning to focus on the unknown samples. However, they fail to distinguish noise from the novel and learnable regularities. The second group of recurring principles include generating intrinsic curious rewards based on error reduction [Schmidhuber (1991b); Oudeyer and Kaplan (2004)], information gain [Storck et al. (1995)], or compression improvement [Schmidhuber (2006)]. This group of principles effectively addresses the above mentioned problem. They are able to guide learning to focus on easily learnable regularities and at the same time filter out noise. The second group of principles forms the basis of Schmidhuber's theory of artificial curiosity, which shows success in speeding up learning and building unsupervised developmental systems [Schmidhuber (1991b)]. Also, Schmidhuber believes that these principles make machines "creative" and intrinsically motivate machines to create action sequences that make data interesting, which forms the basis of his theory of artificial creativity [Schmidhuber (2006)].

# CHAPTER 4 A NOVEL GENERIC COMPUTATIONAL MODEL OF CURIOSITY

In this chapter, we present a computational model of curiosity for intelligent agents. This computational model consists of abstract functions and their interactions between each other, which is inspired by Wooldridge's work [Wooldridge (2002)]. Representing agent models with abstract functions makes them general enough to allow different implementations in different application contexts.

The proposed computational model of curiosity for intelligent agents is built based on the model of a standard agent [Wooldridge (2002)], which is a system situated within an environment and consists of basic functions that sense the environment and act on it to achieve goals. However, to support the mechanism of curiosity, a standard agent is required to go beyond the basic functions and possess other important functions such as memory and learning. Memory stores an agent's previous experiences that are the basis for evaluating the novelty of new experiences. Learning allows an agent to improve its model of the world and make better predictions of the future outcomes, where the accuracy of predictions is the basis for evaluating the surprisingness of new experiences. According to Berlyne's theory [Berlyne (1960b)], both novelty and surprisingness are important collative variables that govern the curiosity arousal mechanism in human beings. Based on these functions, curious functions are introduced by transposing Berlyne's theory [Berlyne (1960b)] and Wundt's theory [Wundt (1874)] from psychology. Curious functions are the meta-level decision-making functions that regulate other functions such as learning and acting of the agent. Next, we will introduce the functions in the proposed computational model of curiosity for intelligent agents and their interactions between each other. We start with the model of a standard agent and then present the two important functions that are required to support curiosity: memory and learning. After that, we will introduce the curious functions in detail.

## 4.1. Standard Agent

This section presents the basic functions that form a standard agent. The description of a standard agent begins with the environment with which the agent interacts. The internal components of a standard agent consist of a set of internal functions that map from one type of internal states to another type of internal states.

### 4.1.1. Environment

According to the classical definition given by Franklin and Graesser [Franklin and Graesser (1996)], an agent is a system situated within and a part of an environment that senses that environment and acts on it, over time, in pursuit of its own agenda and so as to effect what it senses in the future. Hence, an agent must reside in certain environment and act upon it to achieve goals. An agent's environment can be characterized as a set of possible environmental states, denoted by:

$$E \ =< S >$$

where

$$S = \{s_1, s_2, \dots, s_n\}$$

Elements in the angle brackets '<>' represent different types of components that constitute the composite factor that appearing before '=' (this format will be followed throughout the book). In this case, only the environmental state $S$ is highlighted for the environment $E$, which means agents only pay attention to environmental states (e.g., reinforcement learning agents). In other cases when agents also pay attention to other components of the environment, such as objects (e.g., exploratory robots), the environment can be denoted by $E \ =< S, O >$, where $S$ represents a set of environmental states and $O$ represents a set of objects. This can be customized for different types of agents. In this chapter, we focus on the environmental state $S$ for illustration.

## 4.1.2. Internal Functions

To interact with the environment and achieve goals, a standard agent consists of internal functions that observe, deliberate, and then act upon the environment. In this section, the basic internal functions that form a standard agent will be presented.

An agent's ability to effect changes in the environment is determined by the range of actions that it can perform, denoted by:

$$A = \{a_1, a_2, ..., a_m\}$$

Consequently, an agent can be viewed as an abstract function:

$$\boldsymbol{AGT}: S \rightarrow A$$

which maps environmental states to actions.

The behavior of an environment with respect to an agent's actions can be modeled as a function:

$$\boldsymbol{E}: S \times A \rightarrow \varphi(S)$$

which takes the current environmental state $s \in S$ and an agent action $a \in A$, and maps them to a set of environmental states $\varphi(S)$. If all the sets in the range of $\varphi(S)$ are singletons (i.e., the result of performing any action in any state is a set containing a single member that belongs to $S$), then the environment is deterministic, which means its behavior can be accurately predicted.

The abstract function of an agent $\boldsymbol{AGT}$ can be decomposed into a set of internal functions, which will be discussed next.

### 4.1.2.1. Perceiving

The function of perceiving captures an agent's ability to understand information about its environment. The perceiving function can be implemented in hardware agents (e.g., mobile robots) by a video camera or an infra-red sensor, or in software agents (e.g., email assistants) by system commands. An agent's ability to perceive the environment is characterized by a set of percepts:

$$P = \{p_1, p_2, \dots, p_l\}$$

Consequently, the perceiving function is described by:

$$\boldsymbol{P}: S \to P$$

which maps environmental states to the internal percepts of the agent.

### 4.1.2.2. Decision-Making

The function of decision-making encompasses all of the high-level functions of an agent, including belief revision, goal setting, and plan setting. An agent's ability to take appropriate actions is determined by its ability to make decisions, which can be characterized by a set of decision states:

$$D = \{d_1, d_2, \dots, d_k\}$$

Consequently, an agent's decision-making process can be represented by a function:

$$\boldsymbol{D}: P \to D$$

which maps percepts to decision states.

### 4.1.2.3. Acting

Acting is the process of translating high-level goals and plans lower-level commands that can be carried out by effectors. The acting process can be represented by a function:

$$A: D \to A$$

which maps decision states to actions.

### 4.1.2.4. An abstract agent

With all the internal functions described above, a standard agent can be represented abstractly as a compound function:

$$AGT: S \xrightarrow{P} P \xrightarrow{D} D \xrightarrow{A} A$$

$$AGT: A(D(P(S)))$$

The architecture for a standard agent is illustrated in Figure 4.1. The functions described above are illustrated by circular nodes. Solid arrows represent the flow of state variables between functions.

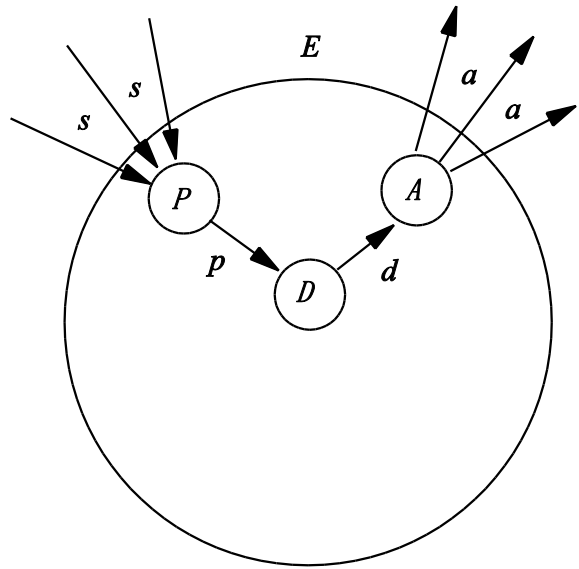

**Figure 4.1 The architecture of a standard agent.**

## 4.2. Memory and Learning

The standard agent described above is a simple reactive agent that acts only on the basis of the current percept, ignoring the rest of the percept history. This type of agent only succeeds when the environment is fully observable. However, in most cases, the environment is only partially observable. For example, a robot's range of perception is limited by the sensors that it possesses. To tackle this problem, an agent usually maintains a model of the environment based on the previous perceptions. This model of the environment describes the part of the world that cannot be seen. However, the initial model may not always succeed when the environment reveals more of itself in front of the agent. This phenomenon requires the agent to be adaptive in order to become more competent than its initial knowledge alone may allow.

These requirements are also the important basis for an agent to become curious. For example, being able to remember past experiences allows the agent to evaluate the novelty of newly encountered stimulus; being able to adapt its initial model of the world allows the agent to make more accurate predictions of the future outcomes, which are important basis for evaluating the surprisingness of new experiences. Here, both novelty and surprisingness are key collative variables that govern the curiosity arousal process [Berlyne (1960b)]. All these requirements point to two important abilities that are desirable by intelligent agents: memory and learning. Next, we will present these two functions and their interactions with other internal functions of an agent.

### 4.2.1. Memory

Memory stores representations of previous experiences, including a number of percepts, decision states, and actions, denoted by:

$$M = \{P^*, D^*, A^*\}$$

where $P^*$, $D^*$, and $A^*$ represent the previous percepts, decision states, and actions, respectively.

These previous experiences, if organized structurally (e.g., neural networks, plan structures, concept maps, etc.), can form the agent's model of the environment. This model of the environment helps the belief revision, goal setting, and plan setting in the agent's decision-making process, which can be characterized by the following function:

$$\boldsymbol{D}_M: P \times M \to D$$

An agent with memory is illustrated in Figure 4.2. In this figure, * refers to the previously experienced percepts, decisions, and actions. It can be seen that memory is located between the processes of perceiving, decision-making and acting.

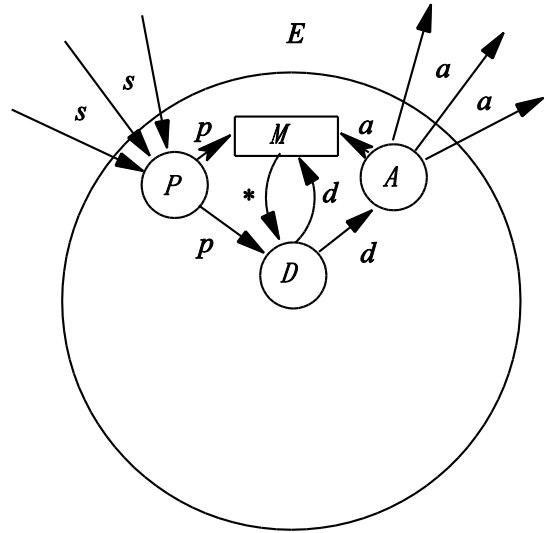

**Figure 4.2 The architecture of an agent with memory.**

### 4.2.2. Learning

Learning enables the agents to initially operate in unknown environments and become more competent than its initial knowledge may allow. The agent's ability to adapt is determined by its ability to update the model of environment stored in its memory and the high level functions in the decision-making process, which is represented by a set of learning rules (or states):

$$L = \{l_1, l_2, \dots, l_i\}$$

The learning rules are triggered depending on the current percepts and the previous decision states stored in memory. For example, prediction errors trigger learning, which are obtained by the differences between the predictions (previous decision states) and the true facts (current percepts). This process can be represented by the following function:

$$\boldsymbol{L}: P \times M \rightarrow L$$

Once the learning rules are triggered (e.g., by newly emerged percepts), they can update the model of environment stored in the agent's memory. For example, a planning agent will incorporate a newly encountered event into its plan structure. This process can be represented by the following function:

$$\boldsymbol{M}: M \times L \to M'$$

where $M'$ is the updated memory.

When the agent's decision is observed to be not optimal, the agent will trigger the learning rules that update the high level functions in the decision-making process to enhance its decision-making ability. For example, a trend analysis agent implemented by neural networks should update its node selection rules when its prediction differs significantly from the true facts. This process can be represented by the following function:

$$\boldsymbol{D}_L: P \times M \times L \times D \to D$$

which requires the current percepts and the previous decision states stored in memory to trigger learning rules and update the decision-making process.

A learning agent is illustrated in Figure 4.3. It can be seen that the learning function takes the outputs of the perceiving function and the memory function as its inputs. The outputs of the learning function in turn influence the memory function and decision-making function.

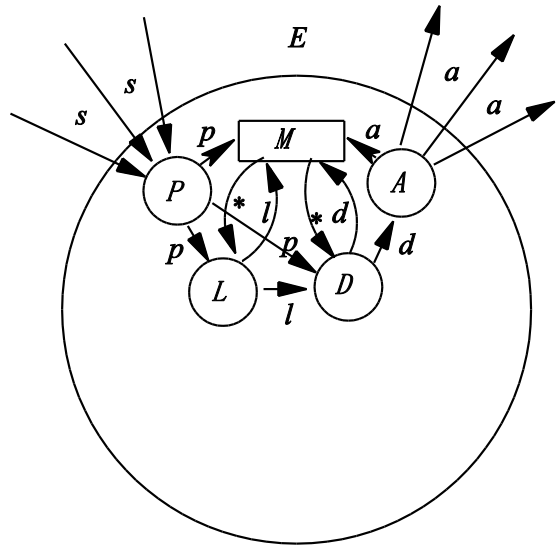

**Figure 4.3 The architecture of a learning agent.**

## 4.3. Curious Functions

Curious functions model the agent's curiosity appraisal process. In order to simulate a human-like curiosity appraisal process, our work is based on Berlyne's [Berlyne (1960b)] and Wundt's [Wundt (1874)] theories in psychology. According to Berlyne, curiosity is a process of stimulus selection, which is governed by a set of collative variables, such as novelty, surprise, uncertainty, conflict, and complexity. Wundt states that the level of curiosity simulation is closely related to two other emotions: boredom and anxiety. According to Wundt, only optimal stimulation level leads to curiosity whereas too less stimulation results in boredom and too much stimulation results in anxiety. Based on these theories, we derived a two-step process model of curiosity arousal, which has been discussed at the end of Chapter2. Following the two-step process model, we propose two curious functions for intelligent agents: stimulus detection and interest evaluation. Stimulus detection is based on Berlyne's theory and corresponds to the first step in the two-step process model, i.e., evaluation of the stimulation level based on collative variables. Interest evaluation is based on Wundt's theory and corresponds to the second step in the two-step process model, i.e., evaluation of the interest level based on the principle of intermediate arousal potential. Next, we will introduce the two curious functions and then discuss their interactions with other internal functions of an agent in detail.

### 4.3.1. Stimulus Detection

According to Berlyne's theory [Berlyne (1960b)], curiosity can be viewed as a process of stimulus selection. The stimuli are characterized by a set of collative variables, which can be represented as follows:

$$V = \{\mathcal{N}, \mathcal{S}, \mathcal{U}, \mathcal{F}, \mathcal{H}, \mathcal{P}\}$$

where $\mathcal{N}, \mathcal{S}, \mathcal{U}, \mathcal{F}, \mathcal{H}$, and $\mathcal{P}$ represent n̲ovelty, s̲urprise, u̲ncertainty, con̲flict, c̲hange, and com̲plexity, respectively. The detection of these collative variables relies on several internal functions of the agent, including perceiving, memory, and decision-making. For example, novelty detection requires a comparison between the current stimuli (obtained by the perceiving function) and previous experiences (stored in memory). Surprise detection involves a comparison between the agent's prediction (generated by the decision-making function) and the true facts (obtained by the perceiving function). Uncertainty is triggered when a stimulus is difficult to classify (by the perceiving function). Conflict occurs when a stimulus triggers multiple decision states (by the decision-making function). Change happens when the state of a stimulus changes (observed by the perceiving function). Complexity is judged by the agent's perceiving function of how much variety or diversity in the stimulus pattern. In summary, the stimulus detection process can be characterized by an abstract function as follows:

$$\boldsymbol{S}: P \times M \times D \to V$$

Note that it is not necessary to model a complete set of collative variables for an agent to be curious, as each collative variable can stand alone to trigger a person's curiosity. Hence, a subset of collative variables can always be chosen according to the agent's functional requirements.

### 4.3.2. Interest Evaluation

According to Wundt's theory [Wundt (1874)], the arousal of curiosity depends on the appropriate level of stimulation that can be induced by a stimulus. Curiosity arouses only if the stimulation is optimal, whereas too little stimulation results in boredom and too much stimulation results in anxiety. The set of emotions closely related to curiosity is represented by:

$$E_m = \{\mathcal{A}, \mathcal{C}, \mathcal{B}\}$$

where $\mathcal{A}$, $\mathcal{C}$, and $\mathcal{B}$ represent anxiety, curiosity, and boredom, respectively.

The process of interest evaluation can be represented by an abstract function as follows:

$$I: \delta(V) \rightarrow E_m$$

where $\delta(V)$ returns the stimulation level induced by the collative variables, and the function $I$ maps the curiosity stimulation level to the interest level (indicated by emotions) based on the Wundt curve (Figure 2.1).

### 4.3.3. Meta-level Decision-making

The two curious functions, i.e., stimulus detection and interest evaluation, are meta-level decision-making functions that interact with agents' learning function and decision-making function to enhance their performances.

Stimulus detection identifies a set of collative variables, all of which reflect certain knowledge gaps between the agent and the environment, which form the motivation for the agents to learn [Loewenstein (1994)]. Hence, the stimulus detection function outputs collative variables that can guide the learning function of an agent to improve its model of the environment. This process can be characterized by the following functions:

$$L_V : V \times L \to L$$

The stimulus detection function can also influence the agent's decision-making function based on different collative variables identified. For example, human beings will have different coping strategies when facing with novelty and conflict. Novelty often triggers a human being to observe the stimulus in order to understand it, whereas conflict often triggers a human being to think of some ideas to resolve the conflict. This process can be characterized by the following functions:

$$D_V : V \times D \to D$$

Interest evaluation determines the agent's emotion states based on the level of stimulation. Emotion states often influences a person's learning ability. For example, a human being will refuse to learn when he/she is bored but learns much faster when he/she is curious. Hence, emotions can influence the agent's learning as follows:

$$L_{E_m} : E_m \times L \to L$$

Emotions also influence a person's decision-making ability. For example, during a study in gambling decisions and job selection decisions, unhappy subjects were found to prefer high-risk/high-reward options unlike anxious subjects who preferred low-risk/low-reward options [Raghunathan and Pham (1999)]. Hence, emotions can influence the agent's decision-making as follows:

$$D_{E_m} : E_m \times D \to D$$

Curious agent architecture is illustrated in Figure 4.4. The two functions: $S$ and $I$ highlighted with dashed box are the curious functions. It can be observed that stimulus detection requires the outputs of perceiving, memory, and decision-making functions, which outputs detected collative variables to trigger interest evaluation and influence the agent's learning and decision-making functions. The emotions generated through interest evaluation also influence the agent's learning and decision-making functions.

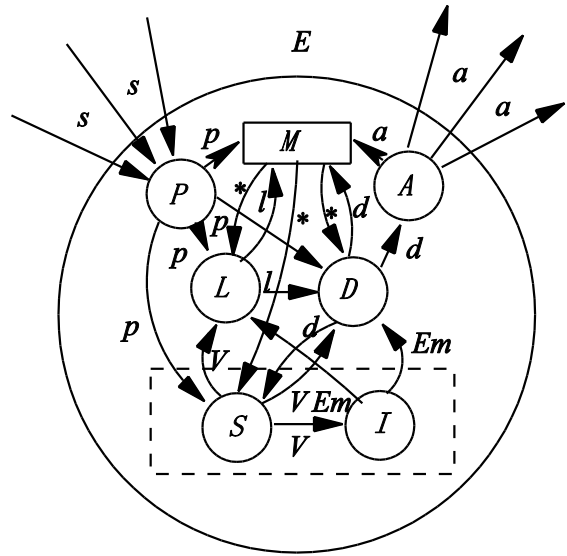

**Figure 4.4 The architecture of a curious agent.**

## 4.4. Summary

In this chapter, we presented a generic computational model of curiosity for intelligent agents. This computational model consists of abstract functions and their interactions between each other, which allow different implementations in different types of agents. This computational model of curiosity is built based on the model of a standard agent, with two important functions introduced that are required to support curiosity appraisal: memory and learning. Based on these functions, two curious functions, i.e., stimulus detection and interest evaluation are proposed based on Berlyne's and Wundt's theories from psychology. The curious functions serve as meta-level decision-making functions that enhance an agent's learning ability and decision-making ability.

# CHAPTER 5 PROMISING APPLICATIONS FOR COMPUTATIONAL CURIOSITY

## 5.1. Machine Learning

The close-knit relationship between curiosity and human learning has inspired many researchers to devise computational forms of curiosity for machine learning systems, with the expectation to enhance learning capability and potentially drive them to evolve into autonomous intelligent agents. The study of computational curiosity in machine learning systems has some overlap with other concepts such as "active learning" and "intrinsically motivated learning".

Active learning, as the name suggests, attempts to machines "active" by allowing the learning system to select actions or make queries that influence what data to be added into its training set [Cohn et al. (1996)]. Active learning is especially useful when data are expensive or difficult to obtain. Curiosity can play a critical role in active learning by helping the system to determine which data are interesting. For example, Scott and Markovitch [Scott and Markovitch (1989)] introduced a curiosity drive into supervised learning systems to actively select the most informative samples for learning. Here, the system is continually directed towards regions with highest uncertainty, which is also a general principle followed by many other active learning algorithms [Fedorov (1972)]. Uğur et al. [Uğur et al. (2007)] infused curiosity into an SVM-based learning system to select interesting training data, which significantly reduces the number of training samples required to learn. Similarly, Pang et al. [Pang et al. (2009)] introduced curiosity into an LDA-based learning system.

Intrinsically motivated learning advocates the development of "intrinsic motivations" for learning systems to achieve task-independent learning [Barto et al. (2004a); Singh et al. (2004)] or autonomous development [Oudeyer and Kaplan (2004)]. These learning approaches are gaining increasing popularity among AI researchers [Baldassarre (2011)]. Intrinsically motivated learning often takes root in a reinforcement learning framework, where intrinsic motivations act as intrinsically generated rewards that are to be maximized. In human psychology, curiosity is known to be one of the most important intrinsic motivations related to learning. Hence, curiosity has often been adopted in intrinsically motivated learning algorithms. For example, Barto et al. [Barto et al. (2004a)] used Berlyne's theory as the psychological foundations to develop his intrinsically motivated learning algorithm. Schmidhuber [Schmidhuber (1991b, 1999, 2009b)] introduced artificial curiosity as intrinsic rewards for general learning algorithms. Oudeyer and Kaplan [Oudeyer and Kaplan (2004)] proposed an intelligent adaptive curiosity mechanism for intrinsically motivated robots.

## 5.2. Robotics

With the attempt to design robots that can autonomously self-develop in a progressive manner, Oudeyer and Kaplan [Oudeyer and Kaplan (2004)] devised for them a mechanism that resembles human curiosity. This mechanism acts as an intrinsic motivation to motivate robots to explore into regions with new knowledge [Oudeyer and Kaplan (2007)], and endows them with the ability to adapt to new environments without prior knowledge or manual adjustment. With a similar goal of designing robots that can self-develop without an explicit teacher, Ngo et al. [Ngo et al. (2012)] applied Schmidhuber's principle of curiosity into a robot arm that enables robots to learn skills through playing. Pape et al. [Pape et al. (2012)] applied the same into a biomimetic robot finger for learning tactile skills.

Traversibility affordance refers to the ability of robots to navigate through an environment with obstacles. This ability is highly dependent on the robot's current location, orientation, and the shape of objects in the environment. In situations where robots physically interact with the environment to explore and learn, assuming traditional reinforcement learning methods are applied, even simple tasks such as avoiding objects may require a large number of trials. This increases the risk of the robot being damaged during the exploration. To address this problem, Uğur et al. [Uğur et al. (2007)] simulated curiosity in robots to select informative training samples, which can significantly reduce the number of interactions required with minimal degradations in the learning process.

Another problem in robotics is self-directed reconfiguration. Reconfigurable robots can rearrange their modules to achieve different structures, behaviors, and functions. Instead of looking into how robots can adapt in an unstructured environment, reconfigurable robots focus on adaptation to changes in their own structures and changes of goals when the actuator or effector of the robot changes. Merrick and Huntington [Merrick and Huntington (2008)] introduced curiosity into reconfigurable robots to select informative samples to learn, so that with fewer interactions, robots can still achieve better learning outcomes.

One of the most important sensory-motor problems for robots when interacting with an environment is to learn a mapping from the gaze space (the location of an object) to the reach space (the movement of arms to grasp the object) in response to changes in the environment (camera replacement or changes in the physical environment) without manual recalibration of the hardware. To address this problem, Karaoguz et al. [Karaoguz et al. (2011)] devised a mechanism of curiosity that drives the exploration into learning spaces where a proper level of complexity is associated with a particular level of capability. With this mechanism, robots can concentrate on highly interesting areas that are neither fully explored nor pure noise. In addition, the mechanism can successfully direct robots' attention to regions where changes have occurred.

Exploration in extreme environments can be a dangerous task for humans. Artificial agents, especially in the form of robots, have been a good substitute for humans to undertake such tasks. Exploration of unknown environments has been an active research field in domains such as planetary exploration, meteorite searches in Antarctic, volcano exploration, and map building of interiors, etc. [Moorehead et al. (2001); Burgard et al. (2002); Macedo and Cardoso (2005)]. In human beings, exploration is often driven by curiosity, a motivating force for attention focus, determination of interest, and gathering knowledge. Based on these observations, researchers devised artificial forms of curiosity for agent to make proper decisions in unknown environments. For example, Macedo and Cardoso [Macedo and Cardoso (2001, 2002, and 2005)] modeled human forms of surprise and curiosity in case-based reasoning frameworks to guide agent's exploration in unknown environments populated with objects. Graziano et al. [Graziano et al. (2011)] also discussed the application of computational curiosity to solve autonomous exploration problems.

To summarize, computational curiosity has the potential to contribute to various aspects of robotic systems, such as selective attention, autonomous learning, and self-direction. Numerous studies on computational curiosity are continuously emerging in this field [Stojanov et al. (2006); Macedo (2010); Oudeyer (2012)].

## 5.3. Artificial Creativity

Computational creativity explores the possibility of machines capable of generating creative artifacts that are commonly defined as being previously unknown, useful and surprising [Boden (2009)]. Computational curiosity has been studied in machines to demonstrate creative behaviors. Based on Csikzentmihalyi's system view of creativity, Saunders [Saunders (2007)] postulated two levels of creativity: the individual level and the society level. According to Saunders, there are two questions to be answered at the individual level of creativity: (1) how to evaluate creativity, and (2) how to produce creativity. Saunders and Gero [Saunders and Gero (2004)] argued that curiosity can be used to guide problem solving by finding interesting design solutions as well as discovering interesting design problems. Curious design agents were proposed to evaluate the interestingness (creativity) of designs based on novelty. Research of curiosity at the society level looks into the socio-cultural influence of curiosity on creativity and the acceptance of creative works by other individuals. Based on previous works, Saunders [Saunders (2011)] studied the society level of creativity by creating a virtual society populated with curious design agents. Simulation results showed that the artificial society exhibited certain similar behaviors as in real human societies.

Schmidhuber [Schmidhuber (2006, 2007, 2009b)] explored the relationship between creativity, artists, humor and fun. He argued that art and creativity can be seen as the by-products of curiosity rewards. He further argued that the optimal curious reward framework can be sufficiently formal and precise to allow the implementation on computers and developmental robots. Schmidhuber [Schmidhuber (2006)] generalized his simple principle of curiosity to form artificial creativity: "the current compressor of a given subjective observer tries to compress his history of acoustic and other inputs where possible" and "the compression progress becomes the wow-effect or intrinsic reward for the 'creative' action selector, which is thus motivated to create more data that allows for more wow-effect". Later, Schmidhuber [Schmidhuber (2013)] proposed a greedy but practical implementation of the basic principle of creativity: POWERPLAY, which can automatically inventor discover problems to train a general problem solver from scratch. In his survey, Schmidhuber [Schmidhuber (2009a)] drew a comparison between his formal theory with less formal works in aesthetics theory and psychology.

## 5.4. Games

With the advances in computer technologies for graphics, processing power, and networking, virtual worlds are emerging as platforms for massive online games. Merrick and Maher [Merrick and Maher (2009)] highlighted the need for new non-player characters to cope with the increasing complexity and functionality of multi-user virtual worlds. They argued that the behavioral traits of humans and animals generated by curiosity can also advance the performance of artificial agents when dealing with complex or dynamic environments, where only limited information is available and the information changes over time. To cope with the difficulty of predefining task specific rules or environment-specific motivation signals, Merrick et al. [Merrick et al. (2008)] introduced a mechanism of curiosity into non-player characters, which enables them to direct attention to relevant information and be curious about changes in the environment. Simulation results showed that the curious motive led the non-player characters to demonstrate higher variety and complexity in behavior patterns [Maher et al. (2008)].

While most of the works that apply computational intelligence to games focused on the generation of behaviors, strategies, or environments, Togelius and Schmidhuber [Togelius and Schmidhuber (2008)] looked at the very heart of games: the rules that define a game. They proposed automatic game designs based on Koster's theory of fun and Schmidhuber's theory of artificial curiosity, where Schmidhuber's theory of curiosity is a coarse approximation of Koster's theory of fun [Koster (2005)], i.e., a game is fun if it is learnable but not trivial.

## 5.5. Affective Computing

Affective computing is computing that relates to, arises from, or deliberately influences emotion or other affective phenomena [Picard (1997)]. It requires multidisciplinary knowledge such as psychology, cognitive science, computer science, and engineering. Affective computing is gaining rapid popularity and has great potential in the next generation of human-computer interfaces. Curiosity is closely related to emotional constructs such as "fear", "pleasure", "boredom", and "anxiety" [Loewenstein (1994)]. Computational curiosity offers a new dimension from which emotions can be appraised, apart from the consequences of events, the actions of agent, and the characteristics of objects [Ortony et al. (1988). The consideration of computational curiosity in affective modeling is especially interesting in learning contexts and social contexts, where curiosity-related emotions significantly influences the believability and performance of emotional agents.

## 5.6. Artificial Companions

Artificial companions, designed to develop a close and long-term human computer relationship, have emerged in the latter half of the 2000s. Two key words, close and long-term, have been guiding the development in this field. Researchers are working on the design of believable human-computer interfaces to provide close interactions (e.g. embodied conversational agent) and robust memory architectures to sustain long-term relationships [Bickmore and Picard (2005); Wilks (2010); Wu et al. (2012b)]. Computational curiosity can be an important dimension to be studied in artificial companions for enhancing both the closeness of interactions and the possibility for long-term relationships. The potential for computational curiosity in creating a closer human-computer relationship draws evidence from psychological findings that curiosity plays an important role in promoting the intimacy of interpersonal relationships in social context. A curious artificial companion can be more responsive; may infuse more novel twists of excitement into interactions, and might induce a natural flow of engagement between the interaction discourses. As for promoting long-term relationships, curiosity can be a motivational force to learn more about the partner and develop a historical knowledge base through interactions. A curious artificial companion may be more interested to know the partner; may be more inquisitive to novel changes of the partner; and may incorporate information of the partner into part of the cognitive development of the companion itself.

## 5.7. Persuasive Technology

Persuasive technology deals with the use of computing systems, devices or applications to gradually change a person's attitudes or behavior [Fogg (2002)]. This technology has the potential to bring constructive changes in health science, safety and education. Examples include a digital doll to persuade kids to eat fruit and vegetables, and a virtual coach to persuade the elderly to exercise more [Fogg (1999)]. Understanding users' curiosity and properly infusing curiosity stimuli into the human-computer interaction process can potentially help intelligent agents achieve persuasive goals. For example, if a sales agent can successfully elicit the customer's curiosity in a product; there will be a higher chance for this product to be sold. Curiosity has been harnessed to "persuade" programmers to increase the correctness in end-user programming [Wilson et al. (2003)].

## 5.8. Agent Negotiation

Negotiation is a process that involves two or more parties to reach an agreement. This mechanism has received increasing attention in multi-agent systems for managing inter-agent dependencies in real time [Jennings and Wooldridge (2002)]. Traditional implementations of negotiation process focused on its rational aspects to build consensus. Recently, [Broekens et al.

(2010)] argued that negotiation is a multifaceted process in which affect plays an important role. Computational curiosity has the potential to influence the human-agent negotiation process by promoting positive emotional states, responsiveness and engagement. Enabling a negotiation agent to understand the curiosity exhibited by a user may allow it to notice the unusual, surprising, or conflicting information offered by the user and reach agreements that are more socially optimal. A negotiation agent that can adapt its decision-making based on the users' curiosity may improve its chance of gaining more utility out of the final agreement.

## 5.9. Trustworthy Computing

Another important issue in multi-agent systems is trust management. It is useful in open and dynamic systems such as peer-to-peer systems, semantic Web, ad hoc networks, and e-commerce, etc. [Ramchurn et al. (2004); Yu et al. (2010, 2012)]. Similar to negotiation, trust management is also closely related to emotion states [Dunn and Schweitzer (2005); Schoorman et al. (2007)]. The motivational role of curiosity in building interpersonal relationships can contribute to the trust building between strangers [Kashdan et al. (2011)]. Computational curiosity can potentially enhance an agent's judgment by making the agent more sensitive to novel, surprising, conflicting, and uncertain information presented in the environment.

# CHAPTER 6 A CURIOUS EXTREME LEARNING MACHINE

## 6.1. Introduction

In this chapter, we focus on realizing the generic computational model of curiosity (Chapter 4) in a type of neural learning agent: an extreme learning machine (ELM) based classifier.

An extremely fast learning neural algorithm referred to as extreme learning machine (ELM) has been developed for single-hidden layer feed-forward networks (SLFNs) by Huang et al. [Huang et al. (2006a,b)]. The essence of ELM is that the hidden layer of SLFNs need not be tuned [Huang et al. (2011)]. ELM randomly assigns hidden neuron parameters and finds the output weights analytically. It has been shown to generate good generalization performance at extremely high learning speed [Huang et al. (2006a, b, c); Liang et al. (2006b)] and has been successfully applied to many real world applications [Huang et al. (2006c); Liang et al. (2006b); Xu et al. (2006); Yeu et al. (2006)].

Although ELM has shown advanced generalization performance with extremely high learning speed, several major issues still remain in ELM:

1. Manually set the number of hidden neurons: The number of hidden neurons needs to be set a priori to training [Feng et al. (2009)]. The number of hidden neurons is usually chosen by trial-and-error.

2. Fixed structure: The network structure is fixed once the number of hidden neurons is set [Rong et al. (2008)]. It cannot evolve, i.e., add or delete hidden neurons, based on the training data.

3. Randomization effect: The random assignment of hidden neuron parameters induces high randomization effect in the generated results.

To address issue 1), several algorithms have been proposed, such as incremental ELM (I-ELM) [Huang and Chen (2007)], enhanced incremental ELM (EI-ELM) [Huang and Chen (2008)], pruning ELM (P-ELM) [Rong et al. (2008)], optimally-pruned ELM(OP-ELM) [Miche et al. (2008)], and error minimized ELM (EM-ELM) [Feng et al. (2009)]. However, all these algorithms can either add neurons (I-ELM, EI-ELM, EM-ELM) or delete neurons (P-ELM, OP-ELM) without being able to adjust network structure based on the incoming data. In other words, they lack the evolving capability. Recently, a meta-cognitive ELM (McELM) has been proposed [Savitha et al. (2014)], which addresses issue 1) and partially issue 2). McELM can decide network structure based on the training data, but it can only add neurons without pruning capability. To our knowledge, few works have been done towards issue 3).

To address all the three issues mentioned above, we propose a curious extreme learning machine (C-ELM) algorithm for classification problems [Wu and Miao (2015)]. It is a psychologically inspired algorithm based on the theory of curiosity [Wu and Miao (2013a)]. In psychology, curiosity is commonly known as the important intrinsic motivation that drives human exploration and learning [Loewenstein (1994)]. The psychological concept of curiosity has been applied in many computational systems to enhance their learning capability (e.g., intrinsically motivated reinforcement learning) [Barto et al. (2004a); Schmidhuber (2009a)] and believability (e.g., curious companions) [Wu et al. (2012a); Wu and Miao (2013b); Wu et al. (2014)]. This is the first attempt to introduce curiosity in an ELM framework.

C-ELM is inspired by the psychological theory of curiosity proposed by Berlyne [Berlyne (1960b)]. Berlyne interpreted curiosity as a process of stimulus selection, i.e., when several conspicuous stimuli are introduced at once, to which stimulus will human respond. He identified several key collative variables, e.g., novelty, uncertainty, conflict, and surprise that govern the stimulus selection process. Based on this theory, C-ELM classifier treats each training data as a stimulus and decides its learning strategy based on the appraisal of collative variables. There are three learning strategies for C-ELM classifier: neuron addition, neuron deletion, and parameter update.

When a new neuron is added, conventional incremental ELM algorithms will randomly assign the center and impact factor of the RBF kernel (other kernels such as linear kernel can also apply) in the new neuron. However, random center selection may require more number of hidden neurons to p-proximate the decision function accurately [Suresh et al. (2008)]. Hence, C-ELM uses data-driven center selection which adopts the current training data that triggers the neuron addition strategy as the center of the new neuron. It removes partially the random effect of the traditional ELM algorithms. Data-driven center selection also allows the class label of the new neuron to be apparent, which enables further analysis of the hidden neurons. During neuron deletion, the most conflicting neuron for the current training data is removed from the network. In literature, various neuron deletion schemes for ELM have been proposed such as pruning based on relevance [Rong et al. (2008)] or based on leave-one-out cross-validation [Miche et al. (2008)]. These techniques, although effective, might render the system slow. Hence, we propose the neuron deletion strategy based on conflict resolution, which helps the system attain fast and efficient convergence. The parameter update is conducted using recursive least squares method.

In the rest of this chapter, we will present the detailed definition of the CELM classifier and evaluate its performance against other popular classifiers on benchmark data sets.

## 6.2. Curious Extreme Learning Machine Classifier (C-ELM)

In this section, we provide a detailed description of the C-ELM classifier. The goal of C-ELM classifier is defined as follows:

**Given**: a stream of training data $\{(\boldsymbol{x}^1, c^1), \dots, (\boldsymbol{x}^t, c^t), \dots\}$, where $\boldsymbol{x}^t = [x_1^t, \dots, x_M^t]$ is a M-dimensional input vector of the $t$th input data, $c^t \in [1, 2, \dots, N]$ is its class label, and $N$ represents the total number of distinct classes. The coded class label $\boldsymbol{y}^t = [y_1^t, ..., y_j^t, ..., y_N^t]$ is obtained by converting the class label $(c^t)$ as follows:

$$y_j^t = \begin{cases} 1, & If\ j = c^t \\ -1, & otherwise \end{cases} j = 1, 2, ..., N$$

**Find**: a decision function $\mathbb{F}$ that maps the input features $(\boldsymbol{x}^t)$ to the coded class labels $(\boldsymbol{y}^t)$, i.e., $\mathbb{F}: \mathfrak{R}^M \to \mathfrak{R}^N$, as close as possible.

To solve this problem, C-ELM employs two major components: an internal cognitive component which is a unified single layer feed-forward neural network (SLFN) and a curiosity appraisal component that consists of curious functions that regulate the extreme learning process.

### 6.2.1. The Internal Cognitive Component: SLFN

The internal cognitive component of the C-ELM is an SLFN with $M$ input neurons, $K$ hidden neurons and $N$ output neurons. For RBF hidden neuron with activation function $g(x): \mathfrak{R} \to \mathfrak{R}$ (e.g., Gaussian), the output of the $k$th hidden neuron with respect to the input $\boldsymbol{x}^t$ is given by:

$$g(\boldsymbol{x}^t, \boldsymbol{a}_k, b_k) = g(b_k ||\boldsymbol{x}^t - \boldsymbol{a}_k||), b_k \in \mathfrak{R}^+$$

where $\boldsymbol{a}_k$ and $b_k$ are the center and impact factor of the kth RBF neuron. $\mathfrak{R}^+$ indicates the set of all positive real values.

The predicted output for the input $\boldsymbol{x}^t$ is given by:

$$\widehat{\boldsymbol{y}}^t = [\hat{y}_1^t, ..., \hat{y}_i^t, ..., \hat{y}_N^t]$$

Here, the output of the $i$th neuron in the output layer is given by:

$$\hat{y}_i^t = \sum_{k=1}^{K} G(\boldsymbol{x}^t, \boldsymbol{a}_k, b_k) \omega_{ki}$$

where $\omega_{ki}$ is the output weight connecting the $k$th hidden neuron to the $i$th output neuron. The output for a chunk of t input data can be written by:

$$\hat{Y} = HW$$

where $H$ is the hidden layer output matrix and $W$ is the weight matrix connecting the hidden neurons to the output neurons as shown below:

$$H = \begin{bmatrix} G(x^1, a_1, b_1) & \cdots & G(x^1, a_k, b_k) \\ \vdots & \ddots & \vdots \\ G(x^t, a_1, b_1) & \cdots & G(x^t, a_K, b_K) \end{bmatrix}_{t \times K}$$

and

$$W = \begin{bmatrix} \omega_{11} & \cdots & \omega_{1n} \\ \vdots & \ddots & \vdots \\ \omega_{K1} & \cdots & \omega_{Kn} \end{bmatrix}_{K \times N}$$

With the cognitive component of C-ELM described above, next, we will introduce the curious functions that regulate the extreme learning process.

### 6.2.2. Curious Functions

C-ELM employs an intrinsically motivated learning paradigm transposed from the psychological theory of curiosity proposed by Berlyne [Berlyne (1960b)]. Learning is regulated based on the curiosity appraisal of input data.

### 6.2.2.1. Stimulus Selection

For each input data, the stimulus selection is governed by four collative variables, i.e., novelty, uncertainty, surprise, and conflict. In this section, we will introduce the definitions of the four collative variables in C-ELM.

**Novelty**: Novelty reflects how much the input data differs from the network's current knowledge. In kernel methods, spherical potential is often used to determine the novelty of data [Subramanian et al. (2013)]. The spherical potential of an input data $x^t$ is defined by (a detailed derivation can be found in [Subramanian et al. (2013)]):

$$\varphi(x^t) = \frac{1}{K} \sum_{k=1}^{K} G(x^t, a_k, b_k)$$

A higher potential indicates that the input data is more similar to the existing knowledge, while a smaller potential indicates that the input data is more novel. Hence, the novelty $\mathcal{N}$ of an input data $x^t$ is determined by:

$$\mathcal{N}(x^t) = 1 - \frac{1}{K} \sum_{k=1}^{K} G(x^t, a_k, b_k)$$

**Uncertainty**: Uncertainty reflects how not confident the network is in its predictions. The confidence of a network is often measured by the posterior probability of the prediction. It has been proven theoretically that hinge-loss function can accurately estimate the posterior probability for a classification problem [Zhang (2004)]. Hence, we use the truncated hinge loss error ($e^t = [e_1^t, \dots, e_j^t, \dots, e_N^t]^T \in \mathfrak{R}^N$) [Suresh et al. (2008)] to measure the prediction error, where each element is defined by:

$$e_j^t = \begin{cases} 0, & if\, \hat{y}_j^t y_j^t > 1 \\ \min(\max(\hat{y}_j^t - y_j^t, -1), 1), & otherwise \end{cases}$$

With the truncated hinge-loss error, the posterior probability of input $x^t$ belonging to class $c$ is given by:

$$p(c|x^t) = \frac{e_j^t + 1}{2}, c = 1,2,\dots,n$$

Since uncertainty measures how not confident a network is in its predictions, we define uncertainty $\mathcal{U}$ of the prediction to an input data $x^t$ by:

$$\mathcal{U}(x^t) = 1 - p(\hat{c}|x^t)$$

where $\hat{c}$ is the predicted class for $x^t$.

Conflict: In psychology, conflict occurs when a stimulus arouses two or more incompatible responses in an organism [Wu and Miao (2013a)]. The degree of conflict depends on the competing strengths of those incompatible responses. For a classifier, conflict can be reflected by the competing strengths of the most _red two output neurons.

Given an input $x^t$, let $\hat{y}_{j1}^t$ and $\hat{y}_{j2}^t$ be the outputs of the most fired output neuron and the second most fired two output neurons, respectively. The more closer $\hat{y}_{j1}^t$ is to $\hat{y}_{j2}^t$, the higher competing strength is between the two output neurons, which indicates a higher conflict between the network's decisions. Hence, the conflict F induced by an input $x^t$ is defined by:

$$\mathcal{F}(x^t) = \begin{cases} 1 - \dfrac{|\hat{y}_{j1}^t - \hat{y}_{j2}^t|}{|\hat{y}_{j1}^t + \hat{y}_{j2}^t|}, & if\ \hat{y}_{j1}^t \hat{y}_{j2}^t > 0 \\ 0, & otherwise \end{cases}$$

**Surprise**: In psychology, surprise indicates a violation of expectation [Wu and Miao (2013a)]. For a classifier, surprise occurs when the predicted output differs from the true class label. The degree of surprise is determined by prediction errors for both the true class and the predicted class. As we adopt hinge-loss error in this work to measure prediction error, the surprise S induced by an input $x^t$ is defined by:

$$\mathcal{S}(x^t) = \begin{cases} |e_c^t \cdot e_{\hat{c}}^t|, & if\ \hat{c} \neq c \\ 0, & otherwise \end{cases}$$

where $\hat{c}$ and $c$ represent the predicted class and the true class, respectively; and $e$ is the hinge-loss error. It can be analyzed that all the four collative variables are within the range of [0, 1].

The collative variables determine the level of curiosity arousal and the learning strategy selection. With the collative variables defined as above, we will introduce the learning strategies in the following section.

### 6.2.2.2. Learning Strategies

C-ELM has three learning strategies: neuron addition, neuron deletion, and parameter update. C-ELM begins with zero hidden neurons, add or delete hidden neurons and update parameters of the existing neurons to achieve an optimal network structure with optimal parameters. The decision on whether to update network structure or update parameters is made based on the appraisal of collative variables. Intuitively, higher values of collative variables induce a higher level of curiosity towards the input data, which require more efforts in learning, i.e., updating network structure, to incorporate the new knowledge; otherwise, simply update parameters of the existing neurons to reinforce the 'familiar' knowledge. Next, we will introduce the three learning strategies of C-ELM in detail.

**Neuron Addition Strategy**: Intuitively, for an input data, if novelty is high and uncertainty is high and surprise is high, it indicates that a misclassification (i.e., surprise high) with high uncertainty in its prediction (i.e., uncertainty high) is caused by the newness of the knowledge (i.e., novelty high). In this case, the network should add new neurons to capture this new knowledge. Hence, given an input $x^t$, the neuron addition condition is:

$$\mathcal{N}(\mathrm{x}^t) > \theta_{N^{add}} \text{ AND } \mathcal{U}(x^t) > \theta_u \text{ AND } \mathcal{S}(\mathrm{x}^t) > \theta_s$$

where $\theta_{N^{add}}$, $\theta_u$, and $\theta_s$ are neuron addition thresholds for novelty, uncertainty, and surprise, respectively. If these parameters are chosen close to 1, then very few input data can trigger the neuron addition strategy and the network cannot approximate the decision function accurately. If these parameters are chosen close to 0, then many input data can trigger the neuron addition strategy, leading to poor generalization ability. In general, $\theta_{N^{add}}$ is chosen in the range of [0.1, 0.5], $\theta_u$ is chosen in the range of [0.1, 0.3], and $\theta_s$ is chosen in the range of [0.2, 0.9].

A typical ELM randomly chooses hidden neuron parameters and finds the output weights analytically. However, random exploration of the feature space may need more hidden neurons to accurately approximate the decision function. In C-ELM, we propose data-driven center selection for the hidden neurons without compromising the extreme learning capability. When a new neuron is added, instead of randomly assigning the center $\boldsymbol{a}_k$, we assign the input features of the current input data as the center, i.e., $\boldsymbol{a}_k = \boldsymbol{x}^t$. Since the center selection is data-driven, we can label the class of the new neuron using the target value of the input data, i.e., $l_k = c^t$. Data-driven center selection allows fast hidden neuron clustering using their class labels and provides class specific information when deleting neurons. The values of the impact factors in the hidden neurons are randomly assigned. Hence, with the new hidden neuron, the dimension of the hidden layer output matrix $\boldsymbol{H}$ increases from $(t-1) \times (k-1)$ to $t \times k$. The target values of the $t$ input data is represented by:

$$Y = \begin{bmatrix} \boldsymbol{y}^1 \\ \vdots \\ \boldsymbol{y}^t \end{bmatrix}_{t \times N}$$

The output weight for $\boldsymbol{W}$ can be analytically found by:

$$\boldsymbol{W} = \boldsymbol{H}^\dagger \boldsymbol{Y}$$

where $\boldsymbol{H}^\dagger$ is the Moore-Penrose generalized inverse of the hidden layer output matrix $\boldsymbol{H}$.

**Neuron Deletion Strategy**: Intuitively, for an input data, if surprise is high and conflict is high and novelty is low, it indicates that a misclassification (i.e., surprise high) occurs for a familiar stimulus (i.e., novelty low) due to high competing strengths between two decisions (i.e., conflict). In this case, the network should adjust its decision-making by strengthening the correct decision and weakening the wrong decision, i.e., deleting the most contributing neuron in the wrong class. Hence, given an input $x^t$, the neuron deletion condition is:

$$\mathcal{S}(x^t) > \theta_\mathcal{S} \text{ AND } \mathcal{F}(x^t) > \theta_\mathcal{F} \text{ AND } \mathcal{N}(x^t) < \theta_{N^{del}}$$

where $\theta_\mathcal{S}$, $\theta_\mathcal{F}$, and $\theta_{N^{del}}$ are neuron deletion thresholds for surprise, conflict, and novelty, respectively. When $\theta_\mathcal{S}$ and $\theta_\mathcal{F}$ are chosen close to 1 and $\theta_{N^{del}}$ is chosen close to 0, then very few input data can trigger neuron deletion, leading to poor generalization ability. When $\theta_\mathcal{S}$ and $\theta_\mathcal{F}$ are chosen close to 0 and $\theta_{N^{del}}$ is chosen close to 1, then many input data can trigger neuron deletion and the network cannot approximate the decision function accurately. In general, $\theta_\mathcal{S}$ is chosen in the range of [0.2, 0.9], $\theta_\mathcal{F}$ is chosen in the range of [0.1, 0.3], and $\theta_{N^{del}}$ is chosen in the range of [0.1, 0.8]. When neuron deletion is triggered, C-ELM will remove the most fired hidden neuron $k$ belonging to the predicted class:

$$\{k | \max(G(x^t, a_x, b_x)) \text{ AND } l_k = c^t\}$$

After the $k$th neuron is removed, the network will re-calculate the output weight $W$ with the $t$ input data.

**Parameter Update Strategy**: When both the neuron addition strategy and the neuron deletion strategy are not triggered, it indicates that the new input data is a 'familiar' one. Hence, the network will update the output weight using recursive least squares to reinforce the familiar knowledge. For the new input data $x^t$, let the partial hidden layer output be represented by $\boldsymbol{h}^t = [G(x^t, a_1, b_1), \ldots, G(x^t, a_K, b_K)]$. The output weights are updated according to [Liang et al. (2006a)] by:

$$\boldsymbol{W}^t = \boldsymbol{W}^{t-1} + \boldsymbol{P}^t \boldsymbol{h}^t ((\boldsymbol{y}^t)^T - (\boldsymbol{h}^t)^T \boldsymbol{W}^{t-1})$$

where

$$P^t = P^{t-1} - \frac{P^{t-1}h^t(h^t)^T P^{t-1}}{1 + (h^t)^T P^{t-1} h^t}$$

### 6.3. Performance Evaluation of C-ELM

The performance of C-ELM is evaluated on the benchmark problems described in Table 6.1 from the UCI machine learning repository, which contains four multi-category classification problems (vehicle classification, iris, wine, and glass identification) and four binary classification problems (liver disorder, PIMA, breast cancer, and ionosphere). The performance of CELM is evaluated in comparison with other popular classifiers such as SVM, ELM and McELM. The results of SVM, ELM and McELM are reproduced from [Savitha et al. (2014)]. For simulating the results of C-ELM, MATLAB 2014b with 3.2 GHz and 16 GB ram was used. The parameters were optimized using grid search.

**Table 6.1 The specification of benchmark datasets on classification problems**

| Data set | # Features | # Classes | # Training data | # Testing data |
|---|---|---|---|---|
| Vehicle | 18 | 4 | 424 | 422 |
| Iris | 4 | 4 | 45 | 105 |
| Wine | 13 | 3 | 60 | 118 |
| Glass identification | 9 | 6 | 109 | 105 |
| Liver disorder | 6 | 2 | 200 | 145 |
| PIMA | 8 | 2 | 400 | 368 |
| Breast cancer | 9 | 2 | 300 | 383 |
| Ionosphere | 34 | 2 | 100 | 251 |

### 6.3.1. Performance Measures

The performance of C-ELM is measured against other popular classifiers using two types of performance measures: average classification accuracy and overall classification accuracy. When the number of samples in each class is highly unbalanced, the average classification accuracy tends to yield more useful information.

**Average classification accuracy**: The average classification accuracy $\eta_a$ is defined by:

$$\eta_a = \frac{1}{C}\sum_i^C \frac{q_i}{N_i} \times 100$$

where $q_i$ is the number of data in class $i$ that have been correctly classified, and $N_i$ is the total number of data in class $i$. It reflects the average ratio of correctly classified data in each class.

**Overall classification accuracy**: The overall classification accuracy $\eta_o$ is defined by:

$$\eta_o = \frac{\sum_i^c q_i}{N_T} \times 100$$

where $N_T$ is the total number of data in the testing data set. It reflects the overall ratio of correctly classified data in the whole testing data set.

### 6.3.2. Performance Study on Multi-category Classification Problems

The performance of the C-ELM on multi-category benchmark classification problems is shown in Table 6.2. It can be observed from Table 6.2 that the generalization performance of C-ELM is better than other classifiers used for comparison on all the multi-category classification problems. Also, the number of hidden neurons added during the evolving process is comparable with other algorithms. For example, C-ELM selected 140 hidden neurons with 6 times neuron deletion for Vehicle classification problem, 6 hidden neurons for Iris problem, 8 hidden neurons for Wine problem, and 52 hidden neurons with 8 times neuron deletion for Glass identification problem. Another advantage of C-ELM in comparison with other self-regulated learning algorithms such as McELM is that it takes a substantially small amount of time for training. For example, McELM takes 40 seconds to train the Vehicle classification problem whereas C-ELM only takes 15.2 seconds. Hence, it shows that C-ELM achieves better performance than other classifiers on multi-category classification problems due to the intrinsically motivated learning mechanism of curiosity.

**Table 6.2 Performance comparison between C-ELM and other popular classifiers on the multi-category classification problems.**

| Data set | Classifier | # hidden neurons | Testing | |
| --- | --- | --- | --- | --- |
| | | | $\eta_o$ | $\eta_a$ |
| Vehicle classification | SVM | 340[a] | 70.62 | 68.51 |
| | ELM | 150 | 77.01 | 77.59 |
| | McELM | 120 | 81.04 | 81.30 |
| | C-ELM[b] | 140 | 81.99 | 82.42 |
| Iris | SVM | 13[a] | 96.19 | 96.19 |
| | ELM | 10 | 96.19 | 96.19 |
| | McELM | 6 | 98.10 | 98.10 |
| | C-ELM[c] | 6 | 99.05 | 99.05 |
| Wine | SVM | 13[a] | 97.46 | 98.04 |
| | ELM | 10 | 97.46 | 98.04 |
| | McELM | 9 | 98.31 | 98.69 |
| | C-ELM[d] | 8 | 99.15 | 99.35 |
| Glass identification | SVM | 183[a] | 70.47 | 75.61 |
| | ELM | 80 | 81.31 | 87.43 |
| | McELM | 72 | 82.86 | 87.40 |
| | C-ELM[e] | 54 | 80.95 | 90.81 |

# support vectors. Number of neuron deleted: (b) 6 (c) 0 (d) 0 (e) 8.

### 6.3.3. Performance Study on Binary Classification Problems

The performance of C-ELM on three binary classification problems is shown in Table 6.3. Table 6.3 shows that C-ELM achieves better generalization performance than other classifiers used for comparison on all the binary classification problems. Also, the total number of hidden neurons added during the evolving process is comparable with other algorithms. For example, C-ELM requires 31 hidden neurons for Liver disorder problem, 33 hidden neurons for PIMA problem, 9 hidden neurons for Breast cancer problem, and 17 hidden neurons for the Ionosphere problem. For binary classification problems, the training time of C-ELM is comparable with other self-regulated learning algorithms such as McELM. For example, it requires 0.73 seconds for C-ELM and 0.95 seconds for McELM to train the Liver disorder problem. Hence, it shows that C-ELM achieves better generalization ability than other classifiers on binary classification problems without compromising the extreme leaning ability of ELM, due to the intrinsically motivated learning mechanism of curiosity.

Table 6.3 Performance comparison between C-ELM and other popular classifiers on the binary classification

| Data set | Classifier | # hidden neurons | Testing | |
|---|---|---|---|---|
| | | | $\eta_o$ | $\eta_a$ |
| Liver disorder | SVM | 141[a] | 71.03 | 70.21 |
| | ELM | 100 | 72.41 | 71.41 |
| | McELM | 50 | 74.48 | 73.83 |
| | C-ELM[b] | 31 | 76.55 | 76.50 |
| PIMA | SVM | 221[a] | 77.45 | 76.33 |
| | ELM | 400 | 76.63 | 75.25 |
| | McELM | 25 | 80.43 | 78.49 |
| | C-ELM[c] | 33 | 81.25 | 80.31 |
| Breast cancer | SVM | 24[a] | 96.60 | 97.06 |
| | ELM | 66 | 96.36 | 96.50 |
| | McELM | 10 | 97.39 | 97.84 |
| | C-ELM[d] | 9 | 97.65 | 98.04 |
| Ionosphere | SVM | 43[a] | 91.24 | 88.51 |
| | ELM | 32 | 89.64 | 87.52 |
| | McELM | 18 | 94.82 | 93.76 |
| | C-ELM[e] | 17 | 95.22 | 95.54 |

# support vectors. Number of neuron deleted: (b) 0 (c) 0 (d) 0 (e) 0.

## 6.4. Summary

In this chapter, we have presented a curious extreme learning machine (CELM) classifier, which is a neural learning agent, based on the generic computational model of curiosity. C-ELM treats each input data as a stimulus for curiosity and performs curiosity appraisal towards each input data based on four collative variables: novelty, uncertainty, conflict, and surprise. Three learning strategies can be chosen from based on the curiosity appraisal results, including neuron addition, neuron deletion, and parameter update. C-ELM enhances traditional ELM algorithms with the evolving capability, which determines optimal network structure dynamically based on the training data. Also, C-ELM reduces partially the random effect of the traditional ELM algorithm by selecting RBF centers based on data instead of random assignment. Moreover, C-ELM employs a novel neuron deletion strategy which is based on conflict resolution. Empirical study of C-ELM shows that the proposed approach leads to compact network structures and generates better generalization performance with fast response, comparing with traditional ELM and other popular classifiers.

# CHAPTER 7 A CURIOUS RECOMMENDER AGENT

## 7.1. Introduction

In this chapter, we focus on the domain of recommender systems and present a curiosity-driven recommendation algorithm that realizes the generic computational model of curiosity (Chapter 4) in another type of intelligent agents: the social recommender agents.

In the current age of information overload, recommender system has become an indispensable technique for filtering and recommending online information. Due to their effectiveness in helping users filtering through the enormous number of items and in helping enterprisers increasing their sales, recommender systems have been successfully adopted by a number of industrial companies, including but not limited to Amazon, Netflix, Yahoo!News, Apple iTunes, etc.

As highlighted in [Resnick and Varian (1997)], the ultimate goal of a recommender system is to suggest particularly interesting items, in addition to indicating those that should be filtered out. Traditional recommender systems are built based on a general consensus that user preferences reflect their underlying interests. Hence, various collaborative filtering techniques [Sarwar et al. (2000)] have been proposed to discover items that best match users' preferences. However, determining interestingness based on user preferences alone is far from sufficient. According to the psychology of interest [Silvia (2008)], the appraisal of interestingness in human beings is closely related to curiosity. Instead of focusing on a person's preferences, curiosity is more related to the novelty and surprisingness in the environment. In this work, we take a new angle to look at the interestingness of recommendation and introduce a novel dimension of social information into the traditional recommender systems: social curiosity.

In real life, it is a commonly observed phenomenon that a person gets curious about the surprising behaviors of his/her friends. For example, Alice knows that her friend Bob always hates horror movies, and the incidence that Bob gives a high rating to a horror movie would interest Alice. In order to find out why Bob gave this surprising rating, Alice may be driven by curiosity to watch this horror movie. This phenomenon is generally known as social curiosity in human psychology. It is the desire to acquire new information about how other people behave, think, and feel [Renner (2006); Wu and Miao (2013a)]. Based on this theory, an item rated by a user's friends can become a recommendation candidate according to how surprising those friends' ratings are for this item.

Motivated as above, we propose a social curiosity inspired recommender system. On top of user preferences, the interestingness of an item is evaluated based on user curiosity as well. The major contributions of this work are summarized as follows. Firstly, we identify a novel dimension of social information for recommender systems: the social curiosity. Secondly, we build a general and parameter-free model for measuring user curiosity in the social contexts. This model takes into consideration the different responses given by a user to different friends' surprising ratings. We also propose and compare three strategies for evaluating user curiosity when multiple friends give surprising ratings to the same item. Thirdly, the model is comprehensively studied with two large scale real world datasets: Douban and Flixster.

The experiment results show that social curiosity significantly improves recommendation diversity and coverage, while maintaining a sufficient level of accuracy. To the best of our knowledge, this is the first work to explore social information for enhancing recommendation diversity and coverage in recommender systems. In the rest of this chapter, we will implement the proposed computational model of curiosity in recommender agents and conduct experimental studies to analyze the effects brought by curiosity.

# CHAPTER 8 A CURIOUS VIRTUAL PEER LEARNER

## 8.1. Introduction

This chapter focuses on developing a curious peer learner that realizes the generic computational model of curiosity to enhance believability. With the advances in computer graphics, communication technologies and networking, virtual worlds are rapidly becoming part of the educational technology landscape [Wiecha et al. (2010); Wu et al. (2013a)]. Dede [Dede (2009)] suggests that the immersive interfaces offered by virtual worlds can promote learning, by enabling the design of educational experiences that are challenging or even impossible to duplicate in real world. In recent years, the usage of virtual worlds within the educational context is growing quickly. The New Media Consortium (NMC) Annual Survey on Second Life (SL) received 170% increase in response rate between 2007 and 2008. They also found that many of the educators, who earlier used the existing SL, have started creating their own virtual worlds in less than a year's time [Harris and Rea (2009)].

Virtual Singapura is a Virtual Learning Environment (VLE) designed to facilitate the learning of plant transport systems in lower secondary school. It has been employed in various studies, such as design perspectives for learning in VLE, pre-service teachers' perspectives on VLE in science education, productive failure and impact of structure on learning in VLE, slow pedagogy in scenario-based VLE, and what students learn in VLE, etc. [Jacobson et al. (2010); Kennedy-Clark (2011, 2009); Kennedy-Clark and Thompson (2011); Tanti and Kennedy-Clark (2010)]. Till date, over 500 students in Singapore and over 300 students in Australia have played Virtual Singapura. During the field studies of Virtual Singapura, several issues with learning in VLE have been observed. First, students tend to spend more time exploring the landscape of the virtual world rather than concentrating on the learning content. Second, some low-functioning students studying alone in VLE often get confused or stuck, and require constant guidance from teachers or game designers to move forward.

Based on these observations, we propose a virtual peer learner to reside in VLE and accompany students to learn. The idea is derived from the common educational practice of peer learning, where students learn with and from each other without the immediate intervention of a teacher [Boud et al. (2001)]. Benefits of a peer learner include: a peer learner can present _learning triggers", that are interactions or experiences causing students to try new things or to think in novel ways; bi-directional peer relationships can facilitate professional and personal growth; and tapping into a learner's own experience can be both affirming and motivating [Eisen (2001)]. Hence, a virtual peer learner has the potential to engage students and motivate them to spend more time on the learning content. Also, a virtual peer learner can potentially help low-functioning students to think and learn better in VLE.

A key design issue for such virtual characters is their believability [Johnson et al. (2000)]: they need to give the users an impression of being lifelike and believable, producing behaviors that appear to the users as natural and appropriate. In order to design a virtual peer learner that can emulate a real student and behave naturally in the learning process, a biologically inspired approach is necessary. In human psychology, studies have shown that curiosity is an important motivation that links cues reflecting novelty and challenge with natural behaviors such as exploration, investigation and learning [Kashdan and Steger (2007)]. In Reiss's [Reiss (2000)] 16 basic desires that motivate our actions and shape our personalities, curiosity is defined as "the need to learn". Attempts to incorporate curiosity into Artificial Intelligence find curious machines have advanced behaviors in exploration, autonomous development, creativity and adaptation [Macedo and Cardoso (2005); Merrick (2008b); Saunders (2007); Scott and Markovitch (1993)]. However, as a basic motivation that drives the learning behaviors of human beings [Reiss (2000)], the role of curiosity in a virtual peer learner has not yet been explored. This creates a challenge for introducing curiosity into virtual peer learners and studying its impact on their learning behaviors. In the rest of this chapter, we will tackle this challenge by implementing the proposed computational model of curiosity in virtual peer learners and conducting experimental studies to analyze the effects brought by curiosity.

# CHAPTER 9 A CURIOUS VIRTUAL LEARNING COMPANION

## 9.1. Introduction

In the previous chapter, we presented a virtual peer learner which realizes the generic computational framework of curiosity. Virtual peer learners are background characters to enrich the virtual learning environment, which do not directly interact with users. However, many educational practices are instantiated through the interactions between teachers and students or between peer learners [Brophy and Good (1974); Webb (1989)]. Hence, a virtual learning companion that can interact with the users would be interesting to enhance their learning experiences. In this chapter, we focus on developing a virtual learning companion which realizes the generic computational framework of curiosity to provide more meaningful interactions with the users. Teaching and learning are highly social activities [Kim (2004)]. With the goal to bring a social context into Virtual Learning Environment (VLE), a growing interest has been shown in designing virtual learning companions that adopt a peer metaphor to simulate peer interactions. Virtual learning companions allow human learners to take advantage of the cognitive and affective gains of human peer-mediated learning in a VLE [Kim (2004)].

The general role of a virtual learning companion is illustrated in Figure 9.1. It can be shown from this figure that a human learner is the main actor in a VLE, who acts upon the environment and learns from it. A virtual learning companion performs as a peer who observes the human learner's actions and their effects, i.e., the environmental changes. Based on the observations, the virtual learning companion performs cognitive and affective reasoning to provide appropriate peer-like interactions with the human learner. Curiosity is an emotional motivation related to exploratory behaviors such as learning, investigation, and exploration, which drives human beings to ask questions and explore for answers. According to the information gap theory by Loewenstein [Loewenstein (1994)], curiosity can be viewed as arising when attention becomes focused on a gap in one's knowledge. Hence, modeling human-like curiosity in a virtual learning companion may allow the companion to discover knowledge gaps and ask relevant questions. These questions can add new ingredients into the interactions provided by the companion, which may help human learners notice the weakness in their knowledge structure and motivate them to actively explore the VLE. However, curiosity has not been studied as a key personality trait in existing virtual learning companions, which creates a challenge for introducing this novel characteristic into such agents and studying the impact brought by it. In the rest of this chapter, we will tackle this challenge by implementing the generic computational framework of curiosity in virtual learning companions and conducting field studies with human users to analyze how an agent with curiosity will impact the human users' learning experience in a human-agent interaction context.

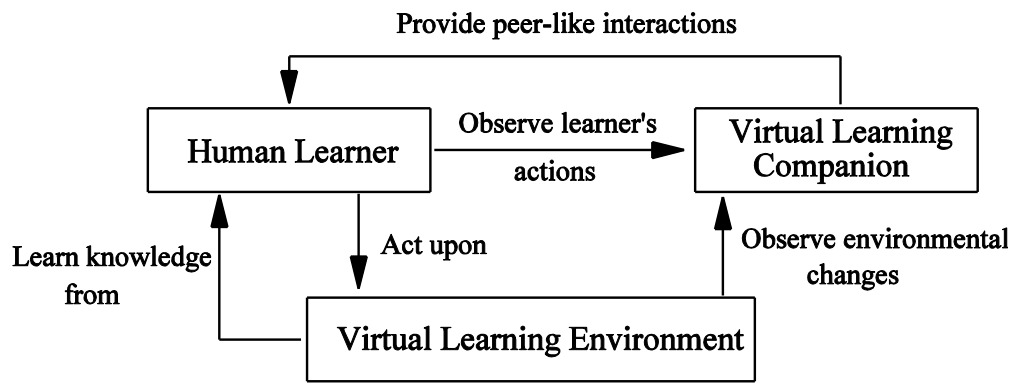

**Figure 9.1 An illustration of the role of a virtual learning companion in VLE.**

# CHAPTER 10 OPEN PROBLEMS

## 10.1. Evaluation of Simulation

Most of the current computational models of curiosity adopt a heuristic approach for the evaluation of stimulations. This is often done based on algorithmic characteristics and goals in machine learning. These models advance the performance of machine learning systems in learning and exploration. However, they have difficulties supporting humanoid agents to evaluate curiosity stimuli in complex environments such as computer-based education, e-commerce, and teleconference. Some models have started from a psychological theory and evaluated stimulation levels based one collative variable or a subset of collative variables. As of yet, how the collative variables affect the level of stimulation, individually or collectively, has not been studied. Both qualitative and quantitative analysis of collative variables can help form a deeper understanding of the working mechanism of computational curiosity, and provide a clear picture of how collative variables are related to performance changes in intelligent agents.

## 10.2. Evaluation of Curiosity

Existing computational models mostly assume a positive correlation between stimulation and curiosity, which may not be always true in human beings. Studies have been done on simulating the Wundt curve to map the level of stimulation to the level of curiosity, but the understanding of how this mapping affects the performance of human-like agents is still unclear. Moreover, in psychology, the curiosity zone is right next to the boredom zone and the anxiety zone. Hence, deeper studies are necessary to provide more proper mapping methods to allow human-like agents to avoid entering the boredom zone or the anxiety zone during exploration.

## 10.3. Interactions between Collative Variables

In the proposed computational model of curiosity, the collative variables are treated as independent factors that stimulate curiosity. This model neglects the inter-influences between the collative variables which have been discussed by Berlyne [Berlyne (1960b)] from the field of psychology. Berlyne postulated that novel stimuli can lead to conflict when a new pattern of stimulus is sufficiently similar to several familiar stimulus patterns, which may induce many incompatible responses. Also, novelty has a connection with complexity. For example, a stimulus with short-term novelty will have a higher degree of temporal complexity than purely repetitive patterns. Alternatively, a higher degree of synchronous complexity may induce a higher degree of relative novelty. Moreover, complexity is also associated with uncertainty and conflict. For example, complex patterns with more parts can assume a larger amount of alternative forms and the degree of uncertainty can be affected by the number of classes in these alternative forms; each sub-element of a complex figure may induce different responses that are incompatible with each other, leading to a higher level of conflict.

Modeling the interactions between the collative variables in the stimulus selection function may yield a more accurate curiosity model for the intelligent agents. However, the interactions between the collative variables are rather vague and complicated. Hence, it is difficult to define the interaction weights using expert knowledge. A possible solution is to learn the interaction weights between the collative variables from data through a machine learning approach.

## 10.4. Neuroscience Inspired Modeling Approaches

The current computational model of curiosity is mainly inspired by the drive theory and optimal arousal theory from the psychological viewpoint. Recently, researchers argued that the psychological models failed to reconcile whether the reward of obtaining new knowledge is mediated through a feeling of _deprivation", as defined by Loewenstein [Loewenstein (1994)], or a feeling of "interest", as defined by Spielberger and Starr [Spielberger and Starr (1994)]. Hence, researchers began to integrate the neurological aspects of reward, wanting, and liking into a new way to understand curiosity, one that is explained by biological processes. Litman [Litman (2005)] developed interest deprivation (I/D) theory of curiosity that incorporates the neuroscience of "wanting" and "liking", which are two systems hypothesized to underlie motivation and affective experience for a broad class of appetites. Litman theory can be summarized in table 10.1. It can be seen that according to this theory, the different states of curiosity are determined by the corresponding levels of the two factors: wanting and liking. This theory motivates the modeling of computational curiosity from a neuroscience inspired approach, which may be considered in the future works to provide a more comprehensive computational model of curiosity.

|  |  | Wanting |  |
| --- | --- | --- | --- |
|  |  | Low level | High level |
| Liking | High Level | Curiosity as a feeling of "interest" (Aesthetic appreciation) | Curiosity as a feeling of "deprivation" (Perceptual/conceptual/fluency) |
|  | Low Level | Ambivalent disinterest or boredom (Spontaneous alternation or novelty seeking) | Need for uncertainty Clarification (Need for cognitive closure; morbid or lurid curiosity) |

**Figure 10.1 An illustration of the interest deprivation (I/D) theory of curiosity.**

## 10.5. Curiosity-based Decision Making

One interesting issue with computational curiosity is the risk management in curiosity-based decision making. As curiosity often leads to explorations, sometimes the agent or human being may be exposed to the possibility of being harmed or causing undesirable consequences to others. Hence, curious machines should operate under the protection of proper risk management systems so that they will not harm themselves. Ethical boundaries should also be defined for curious agents so that they will not intrude on the privacy of the users or other agents.

Koren, Y. (2008). Factorization meets the neighborhood: a multifaceted collaborative filtering model. In proceedings of KDD'08, pages 426_434.

Koster, R. (2005). A theory of fun for game design. Paraglyph Press.

Li, M. and Vitanyi, P. (2008). An introduction to Kolmogorov complexity and its applications. Springer-Verlag New York Inc.

Liang, N., Huang, G., Saratchandran, P., and Sundararajan, N. (2006a). A fast and accurate online sequential learning algorithm for feedforward networks. IEEE Transactions on Neural Networks, 17(6):1411_1423.

Liang, N., Saratchandran, P., Huang, G., and Sundararajan, N. (2006b). Classification of mental tasks from EEG signals using extreme learning machine. International Journal of Neural Systems, 16(01):29_38.

Litman, J. (2005). Curiosity and the pleasures of learning: Wanting and liking new information. Cognition & emotion, 19(6):793_814.

Litman, J. and Spielberger, C. (2003). Measuring epistemic curiosity and its diversive and speci_c components. Journal of Personality Assessment, 80(1):75_86.

Loewenstein, G. (1994). The psychology of curiosity: A review and reinterpretation. Psychological Bulletin, 116(1):75_98.

Ma, H., Zhou, D., Liu, C., Lyu, M., and King, I. (2011). Recommender systems with social regularization. In proceedings of ACM international conference on Web search and data mining, pages 287_296.

Macedo, L. (2010). Selecting information based on artificial forms of selective attention. In proceedings of European Conference on Artificial Intelligence, pages 1053_1054.
86

Oudeyer, P. and Kaplan, F. (2007). What is intrinsic motivation? A typology of computational approaches. Frontiers in Neurorobotics, 1(6):257_262.

Oudeyer, P., Kaplan, F., Hafner, V., and Whyte, A. (2005). The playground experiment: Task-independent development of a curious robot. In proceedings of the AAAI Spring Symposium on Developmental Robotics, pages 42_47.

P. Li, X. Luo, X. Meng, C. Miao, M.a He and X. Guo. A Two- Stage Win–Win Multiattribute Negotiation Model: Optimization and Then Concession. Computational Intelligence, vol. 29, pp.577–626, 2013.

P. Wu, S. C. H. Hoi, H. Xia, P. Zhao, D. Wang, and C. Miao, "Online multimodal deep similarity learning with application to image retrieval," in Proceedings of the 21st ACM International Conference on Multimedia (MM '13) #. ACM, October 2013, pp. 153–162.

P. Wu, Y. Ding, P. Zhao, C. Miao, and S. Hoi, "Learning relative similarity by stochastic dual coordinate ascent," in Proceedings of the 28th AAAI Conference on Artificial Intelligence (AAAI'14). AAAI, July 2014.

Pang, S., Ozawa, S., and Kasabov, N. (2009). Curiosity driven incremental LDA agent active learning. In proceedings of International Joint Conference on Neural Networks, pages 2401_2408.

Pape, L., Oddo, C., Controzzi, M., Cipriani, C., F_'oster, A., Carrozza, M., and Schmidhuber, J. (2012). Learning tactile skills through curious exploration. Frontiers in Neurorobotics, 6:6.

Picard, R. (1997). A_ective computing. The MIT Press.

Raghunathan, R. and Pham, M. T. (1999). All negative moods are not equal: Motivational in_uences of anxiety and sadness on decision making. Organizational behavior and human decision processes, 79(1):56_77.
89

Schmidhuber, J. (1991c). A possibility for implementing curiosity and boredom in model-building neural controllers. In proceedings of the International Conference on Simulation of Adaptive Behavior: From Animals to Animats, pages 222_227.

Schmidhuber, J. (1999). Artificial curiosity based on discovering novel algorithmic predictability through coevolution. In proceedings of the Congress on Evolutionary Computation, pages 1612_1618.

Schmidhuber, J. (2002). Exploring the predictable. Advances in Evolutionary Computing, pages 579_612.

Schmidhuber, J. (2006). Developmental robotics, optimal artificial curiosity, creativity, music, and the _ne arts. Connection Science, 18(2):173_187.

Schmidhuber, J. (2007). Simple algorithmic principles of discovery, subjective beauty, selective attention, curiosity & creativity. In Proc: International Conference on Discovery Science, pages 26_28.

Schmidhuber, J. (2009a). Formal theory of creativity, fun and intrinsic motivation. IEEE Transaction on Autonomous Mental Development, 2(3):230_ 247.

Schmidhuber, J. (2009b). Simple algorithmic theory of subjective beauty, novelty, surprise, interestingness, attention, curiosity, creativity, art, science, music, jokes. Journal of the Society of Instrument and Control Engineers, 48(1):21_32.

Schmidhuber, J. (2013). Powerplay: Training an increasingly general problem solver by continually searching for the simplest still unsolvable problem. Frontiers in psychology, 4(313).

Schmitt, F. and Lahroodi, R. (2008). The epistemic value of curiosity. Educational Theory, 58(2):125_148.